%% file: main.tex
\documentclass[runningheads]{llncs}

\usepackage[mobile]{eccv}

\usepackage[pagebackref=false,breaklinks=true,colorlinks,citecolor=eccvblue,bookmarks=true,bookmarksnumbered=true]{hyperref}
\hypersetup{
  pdftitle={Event-Aided Time-to-Collision Estimation for Autonomous Driving},
  pdfsubject={Robotics, Computer Vision, Autonomous Driving},
  pdfauthor={Jinghang Li, Bangyan Liao, Xiuyuan Lu, Peidong Liu, Shaojie Shen, Yi Zhou},
  pdfkeywords={Event cameras, Time to collision, Autonomous driving}
}
\usepackage[absolute]{textpos}

\usepackage{eccvabbrv}

\usepackage{graphicx}
\usepackage{booktabs}
\usepackage{multirow}
\usepackage{makecell}
\usepackage{threeparttable}
\usepackage{adjustbox}
\usepackage[bottom]{footmisc}
\usepackage[accsupp]{axessibility}

\usepackage{lipsum}

\usepackage{hyperref}

\usepackage{orcidlink}

\usepackage{marvosym}
\usepackage{cancel}

\usepackage{xcolor}
\usepackage{pifont}
\hypersetup{colorlinks=true}
\newcommand{\cmark}{\color{green}\ding{52}}

\newcommand{\xmark}{\color{red}\ding{54}}

\usepackage{siunitx}

\usepackage{pgfplots}
\pgfplotsset{compat=newest}

\input{chapters/math_macros.tex}

\begin{document}

\title{Event-Aided Time-to-Collision Estimation for Autonomous Driving}

\definecolor{somegray}{gray}{0.5}
\newcommand{\darkgrayed}[1]{\textcolor{somegray}{#1}}
\begin{textblock}{11}(2.5, -0.1)  % {hsize}(hpos,vpos)
\begin{center}
\darkgrayed{This paper has been accepted for publication at the European Conference on Computer Vision (ECCV), 2024. \copyright Springer}
\end{center}
\end{textblock}

\titlerunning{Event-Aided Time-to-Collision Estimation for Autonomous Driving}
\authorrunning{J. Li, B. Liao et al.}

\author{
Jinghang Li\inst{1}\textsuperscript{$\star$} \orcidlink{0000-0001-6196-6165} \and
Bangyan Liao\inst{2}\textsuperscript{$\star$}\orcidlink{0009-0007-7739-4879} \and
Xiuyuan Lu\inst{3}\orcidlink{0000-0001-6376-8584} \and
Peidong Liu\inst{2}\orcidlink{0000-0002-9767-6220} \and \\
Shaojie Shen\inst{3}\orcidlink{0000-0002-5573-2909} \and
Yi Zhou\inst{1\text{\Letter}} \orcidlink{0000-0003-3201-8873}
}

\institute{
    School of Robotics, Hunan University\and
    School of Engineering, Westlake University\and
    Dept. of ECE, Hong Kong University of Science and Technology
}

\maketitle
\input{chapters/00_abstract}
\renewcommand{\thefootnote}{\relax}\footnotetext{\textsuperscript{$\star$} equal contribution; \textsuperscript{\Letter} corresponding author (eeyzhou@hnu.edu.cn).}
\renewcommand{\thefootnote}{\arabic{footnote}}
\setcounter{footnote}{0}
\input{chapters/01_introduction}
\input{chapters/02_related_work}
\input{chapters/03_methodology}

\input{chapters/04_experiments}

\input{chapters/05_conclusions}
\input{chapters/05a_acknowledgment}

% The document is based on Springer LNCS instructions as well as on ECCV policies, as established over the years.
\appendix
\section*{SUPPLEMENTARY MATERIAL}
\input{chapters/06_appendix} % The supplementary material goes to main_supp.tex.

\bibliographystyle{splncs04}
\bibliography{reference}
\end{document}

%% file: chapters/math_macros.tex
\DeclareMathOperator*{\argmin}{arg\,min}

\global\long\def\bfa{\mathbf{a}}

\global\long\def\bfe{\mathbf{e}}

\global\long\def\bfn{\mathbf{n}}

\global\long\def\bfp{\mathbf{p}}

\global\long\def\bfu{\mathbf{u}}

\global\long\def\bfx{\mathbf{x}}

\global\long\def\bfP{\mathbf{P}}

\global\long\def\ttB{\mathtt{B}}

\global\long\def\ttT{\mathtt{T}}

\global\long\def\cE{\mathcal{E}}

\global\long\def\cT{\mathcal{T}}

\global\long\def\bfomega{\boldsymbol{\omega}}

\global\long\def\bfnu{\boldsymbol{\nu}}

\global\long\def\tref{t_{\text{ref}}}

\newcommand{\os}[5]
{
{}^{#1}_{#2}{#3}^{#4}_{#5}
}

%% file: chapters/00_abstract.tex
\begin{abstract}
Predicting a potential collision with leading vehicles is an essential functionality of any autonomous/assisted driving system.
One bottleneck of existing vision-based solutions is that their updating rate is limited to the frame rate of standard cameras used. 
In this paper, we present a novel method that estimates the time to collision using a neuromorphic event-based camera, a biologically inspired visual sensor that can sense at exactly the same rate as scene dynamics. 
The core of the proposed algorithm consists of a two-step approach for efficient and accurate geometric model fitting on event data in a coarse-to-fine manner.
The first step is a robust linear solver based on a novel geometric measurement that overcomes the partial observability of event-based normal flow.
The second step further refines the resulting model via a spatio-temporal registration process formulated as a nonlinear optimization problem.
Experiments on both synthetic and real data demonstrate the effectiveness of the proposed method, outperforming other alternative methods in terms of efficiency and accuracy. 
Dataset used in this paper can be found at \url{https://nail-hnu.github.io/EventAidedTTC/}.
\keywords{Event cameras \and Time to collision \and Autonomous driving}
\end{abstract}

%% file: chapters/01_introduction.tex
\section{Introduction}
\label{sec: introduction}

The ability to identify objects that pose a conflicting threat to the host vehicle and make a brake decision is fundamentally important in Level-2 and beyond autonomous/assisted driving techniques.
The time for such a potential collision between the host vehicle and the leading vehicle is coined by \textit{time to collision} (TTC), which has been used for alerting drivers and autonomous driving systems for deceleration.
The most straightforward way to calculate TTC requires knowing the absolute distance and relative speed between the two vehicles.
Thus, LIDAR/Radar based solutions are typically applied \cite{shaw1996vehicle, widman1998development}.
However, it has been witnessed that experienced human drivers perform well on controlling the relative speed adaptively, indicating that visual information alone is enough for the task of TTC estimation.

\input{floats/fig_eye_catcher}
There is a large body of literature on TTC estimation using standard cameras \cite{nelson1989obstacle, meyer1992estimation, meyer1994time, dagan2004forward, negre2008real, souhila2007optical, stabinger2016monocular, poiesi2016detection}.
The most widely studied case uses a monocular standard camera and calculates the TTC using a pair of successive images.
The leading car's scale (i.e., image size) varies in the presence of relative speed w.r.t the host vehicle, and the TTC can be recovered from such a scale variation.
One bottleneck of these solutions is that their updating rate is limited to the frame rate of standard cameras.
Standard cameras used in autonomous/assisted driving systems typically run around 10 Hz due to the consideration of cost, bandwidth, and energy consumption.
The interval between two successive exposures (i.e., 100 ms), even not taking into account the computation time of the applied TTC algorithm, can be a notably big latency to a collision warning system, especially when the relative speed increases drastically.

Event-based cameras are biologically-inspired novel sensors that mimic the working principle of vision pathway of human beings.
Different from standard cameras, an event-based camera reports only brightness changes at each pixel location asynchronously.
This unique characteristic leads to much better performance in terms of temporal resolution and dynamic range.
Therefore, event cameras are inherently well-qualified to deal with robotics applications involving aggressive motion \cite{Rebecq17bmvc, falanga2020dynamic} and high dynamic range (HDR) scenarios \cite{Rebecq19pami, Pan19cvpr}.

In this paper, we look into the problem of TTC estimation using an event-based camera.
We focus on predicting a potential collision from event data given the presence of a leading vehicle as a prior.
Our goal is to obtain accurate TTC estimates at an ultra framerate by leveraging the asynchronous characteristic and high temporal resolution of event cameras.
As illustrated in Fig.~\ref{fig:eye catcher}, the image size of the leading vehicle varies when there is a relative speed between the leading vehicle and the observer.
Such a scale variation induces an optical flow field of radial expansion (or contraction), and meanwhile a set of events are triggered as those contours expand (or contract). Ideally, the TTC can be derived from the divergence of this flow field\cite{nelson1989obstacle}.
However, recovering the full flow field from event data is not trivial and even intractable because event data report only photometric variations along local gradient directions.
To this end, we propose a geometric method that can recover the TTC from the profile of event-based normal flow. 
Specifically, the contribution of this paper consists of:
\begin{itemize}
    \item A time-variant affine model, derived from the dynamics of contour points, that accurately guides the warping of events in the spatio-temporal domain.
    \item A robust linear solver based on the proposed geometric measurement that can overcome the partial observability of event-based normal flow.
    \item A nonlinear solver that further refines the resulting model via a spatio-temporal registration process, in which efficient and accurate data association is established by using an ingeniously modified representation of events, i.e., linear time surfaces.
\end{itemize}

\textit{Outline: } The rest of the paper is organized as follows.
Section~\ref{sec: related work} gives a literature review on the development of TTC techniques.
Section~\ref{sec:methodology} discloses the proposed event-based method for TTC estimation.
Our method is evaluated in Sec.~\ref{sec:evaluation} and conclusions are drawn in Sec.~\ref{sec:conclusion}.

%% file: floats/fig_eye_catcher.tex
\begin{figure}[ht]
    \centering
    \begin{subfigure}[t]{0.6\textwidth}
        \centering
        \includegraphics[width=0.95\textwidth]{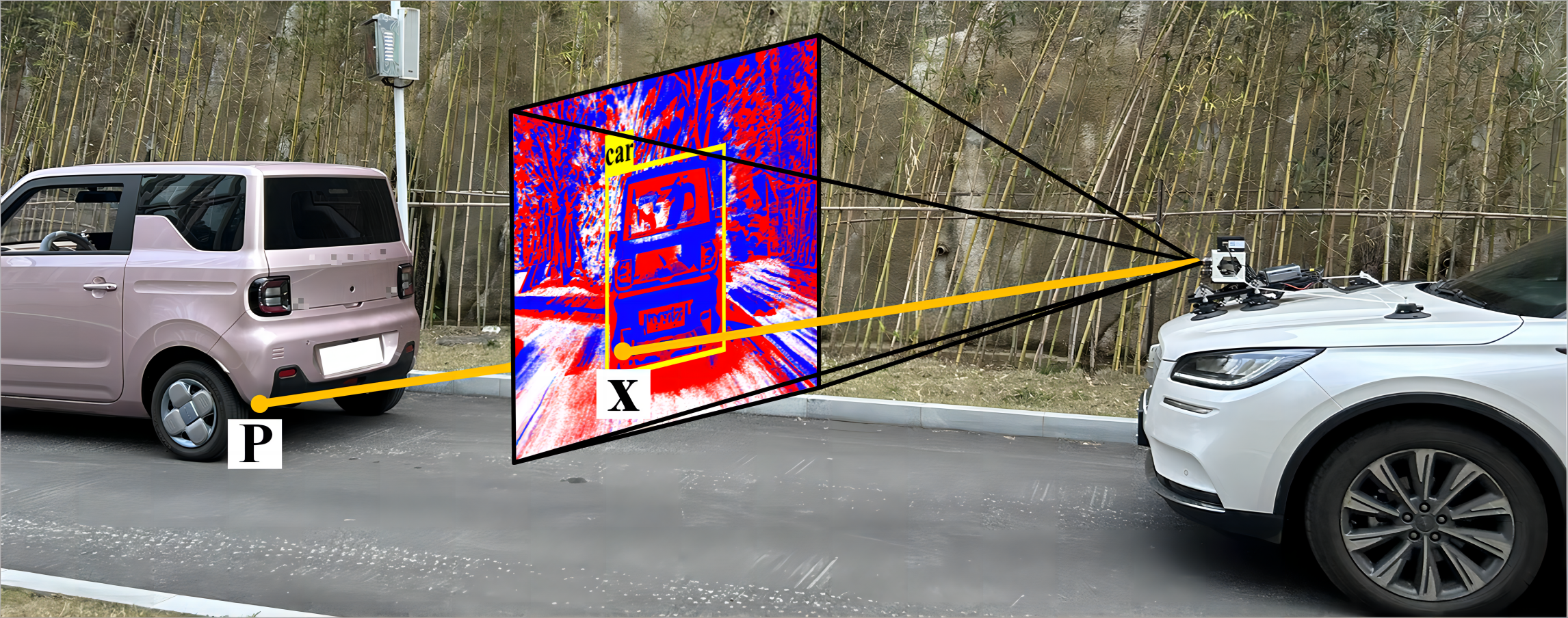}
        \caption{Geometry of the event-based TTC problem.}
        \label{fig:geometry of TTC}
    \end{subfigure}
    \begin{subfigure}[t]{0.35\textwidth}
        \centering
        \includegraphics[width=0.6\textwidth]{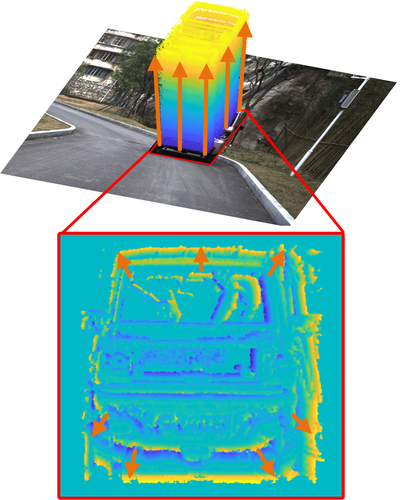}
        \caption{Motion of contours and the corresponding events triggered.}
        \label{fig:generation of events in TTC}
    \end{subfigure}
    \caption{Illustration of the TTC estimation problem.
(a) The leading vehicle is perceived by an following observer. 
(b) Events are triggered as the image size of the leading vehicle varies in the presence of relative speed.
A motion flow field of outward expansion is visible in the close-up of the proposed image representation of events (see \ref{subsec:Model Fitting via Spatio-Temporal Registration} for detail), indicating a decrease in the relative distance and an increase of collision risk.}
    \label{fig:eye catcher}  
\end{figure}

%% file: chapters/02_related_work.tex
\section{Related Work}
\label{sec: related work}

\subsection{History of TTC Estimation}
\label{subsec: History of TTC Estimation}
The earliest work in regards to TTC estimation can date back to the 1980s.
D.N.Lee claimed that TTC was widely used in nature, such as birds' landing~\cite{lee1980optic}, and he also argued that TTC could provide enough information for a driver to adjust the velocity w.r.t obstacles~\cite{lee1976theory}.
The connection between the TTC and the 2D motion (optical flow) field is first disclosed in \cite{nelson1989obstacle}, which derives the TTC from the motion field's divergence and realizes obstacle avoidance.
Successive work \cite{meyer1992estimation, meyer1994time} prove that the first-order visual models are sufficient to obtain TTC estimates, and \cite{meyer1992estimation} further demonstrates that normal flow is enough to recover the divergence as long as the displacement between the successive frames is small.
Since then, a large number of similar ideas have been witnessed \cite{souhila2007optical, stabinger2016monocular, poiesi2016detection}, including some variants that leverage the scale variation of features \cite{dagan2004forward, negre2008real}.
More recently, learning approaches have been applied to solve the TTC problem.
\cite{manglik2019forecasting} proposes an end-to-end deep learning approach that directly estimates the time to collision using a sequence of images as input. 
\cite{badki2021binary} formulates the TTC problem (on multiple obstacles) as a pixel-level multiple binary classification task for different TTC thresholds. 
Although modern state-of-the-art solutions have pushed a remarkable step forward in terms of computational efficiency, the latency that originates from the sensing end (i.e., frame-based cameras) still exists.
The only way to circumvent this bottleneck is to bring in sensors that have a higher temporal resolution to sense at the same rate as the scene dynamics.

\subsection{Tryouts with Event-based Cameras}
\label{subsec: Tryouts with Event-based Cameras}

The research on the TTC problem, or more generally speaking, obstacle avoidance using event-based vision has received large attention from the community of robotics\cite{Sanket20icra, falanga2020dynamic, walters2021evreflex, rodriguez2022free} and neuromorphic engineering \cite{clady2014asynchronous}.
Inspired by methods using standard cameras, most event-based solutions are built on top of optical flow from event data \cite{benosman2013event, Zhu18rss, liu2018adaptive, Zhu19cvpr, almatrafi2020distance}.
Among the flow-based methods, we find a major category \cite{clady2014asynchronous, dinaux2021faith}, which typically starts with localizing the Focus of Expansion (FOE) by increasingly shrinking the intersection area of multiple negative half-planes. 
These methods facilitate UAV heading and TTC estimation in the presence of marking lines. However, the generalizability of TTC estimation is hindered by the specific conditions it relies on (i.e., requiring marking lines), and its accuracy is notably influenced by the imprecisely estimated FOE.
Besides, it’s also worth mentioning that a parametric model of events’ motion can be regressed from event data without explicitly solving the optical flow problem.
A unifying contrast maximization framework presented in \cite{Gallego18cvpr}, also known as Contrast Maximization (CM), can fit a family of geometric models on event data without explicit data association, such as features and optical flow.
Given the resulting geometric model (e.g., an affine model), the TTC can be easily recovered using the derivation in \cite{nelson1989obstacle, meyer1992estimation, meyer1994time}.
Such pipelines have been witnessed in \cite{shiba2022fast, mcleod2023globally}.
Our method is similar to \cite{mcleod2023globally} in the sense that we also leverage the fact that TTC is the inverse of the radial flow field's divergence, but with differences: 
($i$) Our geometric model is more general and can handle 2D planar translation, while \cite{mcleod2023globally} makes a simplification for competitive running time;
($ii$) Our model fitting method via spatio-temporal registration is more efficient than CM-based pipelines due to the least-squares nature;
($iii$) We provide a robust linear solver for initialization, which locates the convergence basin in a closed-form manner.

%% file: chapters/03_methodology.tex
\section{Methodology}
\label{sec:methodology}
In this section, we first state the specific problem solved in this work (Sec.~\ref{subsec: problem statement}).
Second, we introduce a time-variant geometric model, derived from the real dynamics of contour points, that accurately guides the warping of events (Sec.~\ref{subsec: A Flow-Dynamics Consistent Model}).
Third, we detail the method for geometric model fitting on event data via solving a spatio-temporal registration problem, in which a novel representation of events is proposed for efficient and accurate data association (Sec.~\ref{subsec:Model Fitting via Spatio-Temporal Registration}). 
To initialize, we also propose a robust linear solver, which is based on a novel geometric measurement that overcomes the partial observability of event-based normal flow (Sec.~\ref{subsec:Initialization}).
Finally, the design of a complete forward collision warning (FCW) system built on top of the proposed event-based method is discussed (Sec.~\ref{subsec:FCW system}).
\subsection{Problem Statement}
\label{subsec: problem statement}
Given the identification result of a preceding vehicle as a prior, the goal of this work is to estimate the TTC during the blind period of the standard camera (\ie the time interval between two successive exposures) using event data as input.
The presence of relative speed between the proceeding vehicle and the observer leads to a size variation of the former on the observer's image plane, and such a zoom-in or zoom-out behavior of the car's contours will trigger a set of events.
The TTC information is encoded in the flow field of these contour points, which can be typically parametrized using a geometric model.
Our method estimates the TTC by fitting such a parametric model on event data.
\subsection{A Flow-Dynamics Consistent Geometric Model}
\label{subsec: A Flow-Dynamics Consistent Model}

Consider a proceeding vehicle followed by an observer running in the same lane.
If the relative instantaneous angular velocity and linear velocity between them are denoted in the observer's body frame as $\os{\ttB}{}\bfomega{}{}=[\omega_{x},\omega_{y},\omega_{z}]^{\ttT}$ and $\os{\ttB}{}\bfnu{}{}=[\nu_{x},\nu_{y},\nu_{z}]^{\ttT}$, respectively,
the scene flow of a 3D contour point $\os{\ttB}{}\bfP{}{} = [X,Y,Z]^{\ttT}$ will be
\mbox{$\dot{\os{\ttB}{}\bfP{}{}} = -\os{\ttB}{}\bfomega{}{} \times \os{\ttB}{}\bfP{}{} - \os{\ttB}{}\bfnu{}{}.$}
Let $\bfp = [x,y]^{\ttT} = [X/Z, Y/Z]^{\ttT}$ be the image of $\bfP$ represented by the normalized coordinates.

Usually, the relative angular velocity can be omitted in our context and it's also reasonable to assume a constant relative linear velocity during a short time interval.
Some existing methods \cite{souhila2007optical, stabinger2016monocular, poiesi2016detection} using as input the optical flow (or 2D feature correspondences) between two frames further assume a constant optical flow, and thus, the flow vector from start time $t_0$ to reference time $t_{\tt{ref}}$ becomes
\begin{equation}
\bfp(t_{\tt{ref}}) - \bfp(t_0) = \frac{1}{Z(t_0)} \begin{bmatrix}
-{\nu}_x + x(t_0){\nu}_z \\ 
-{\nu}_y + y(t_0){\nu}_z
\end{bmatrix}(t_{\tt{ref}} - t_0).
\label{eq:constant parametric model}
\end{equation}
This gives rise to a constant parametric flow model, \eg~an affine transformation \cite{stabinger2016monocular}, and the TTC can be simply calculated from the model parameters \cite{jahne1999handbook} (p. 383).
However, such a strong assumption may violate the time-variant nature of the flow field, especially when the relative distance varies notably within a short period of time.
The key to the problem of geometric model fitting on event data is applying an accurate model. Therefore, we propose a time-variant parametric model derived from the real dynamics of the flow field.

Written in a continuous-time form, the optical flow $\bfu$ can be defined as the ordinary differential equation (ODE) of $\bfp$,
\begin{align}
\bfu(t) \doteq \dot{\bfp}(t) = \begin{bmatrix}
\dot{x}(t) \\ 
\dot{y}(t)
\end{bmatrix} = \frac{1}{Z(t)} \begin{bmatrix}
-{\nu}_x + x(t){\nu}_z \\ 
-{\nu}_y + y(t){\nu}_z
\end{bmatrix},
\label{eq:ODE}
\end{align}
based on which the position of $\bfp$ at $t_{\tt{ref}}$ can be accurately obtained by the following integral: 
$\bfp(t_{\tt{ref}}) = \int_{t_0}^{t_{\tt{ref}}} \bfu(t) dt + \bfp(t_0)$.
By solving the above ODE and substituting the boundary condition at $t_{\tt{ref}}$ into its general solution, we obtain
\begin{equation}
\bfp(t_{\tt{ref}}) = \underbrace{\frac{1}{Z(t_{\tt{ref}})} \begin{bmatrix}
-{\nu}_x + x(t_0){\nu}_z \\ 
-{\nu}_y + y(t_0){\nu}_z
\end{bmatrix}}_{\mathcal{A}(\bfp(t_0);~\boldsymbol{\nu}/{Z(t_{\tt{ref}})})}(t_{\tt{ref}} - t_0) + \bfp(t_0)  ,
\label{eq:time-variant geometric model}
\end{equation}
where $\mathcal{A}(\bfp(t_0); \boldsymbol{\nu} / {Z(t_{\tt{ref}})})$ consists of a time-variant affine model and can be regarded as the average flow during the time interval. 
Note that the only difference between Eq.~\ref{eq:time-variant geometric model} and Eq.~\ref{eq:constant parametric model} is replacing $Z(t_0)$ with $Z(t_{ref})$.
This small change leads to a more accurate geometric model, which is crucial to the following spatio-temporal registration.
A justification of this can be found in the supplementary material.

\subsection{Model Fitting via Spatio-Temporal Registration}
\label{subsec:Model Fitting via Spatio-Temporal Registration}
\input{floats/fig_event_representations}

\input{floats/fig_representation_analysis}
Our goal is to recover the above mentioned geometric model through inferring the event data association, namely determining implicitly which of these events are triggered by the same contour point.
Such an operation is typically referred to as spatio-temporal registration~\footnote{Generally speaking, any approach that establishes data association by leveraging information from the spatio-temporal domain falls into this category of methods.
}, which has been widely applied to event-based model fitting tasks, such as 3D rotation estimation \cite{liu2021spatiotemporal}, 6D pose estimation
\cite{zhou2021esvo} and motion segmentation \cite{zhou2021emsgc}.

We borrow the general idea of spatio-temporal registration to the problem of fitting an affine model.
Instead of using Time Surfaces (TS), an image-like representation of event data in \cite{lin2020efficient, zhou2021esvo}, we propose a linear counterpart called Linear Time Surface (LTS).
Different from the ordinary TS, an LTS stores the time difference between reference time $t_{ref}$ and the timestamp of the event triggered closest to $t_{ref}$. 
As is known, each event $\bfe_{k} = (\bfx_{k}, t_k, p_k)$ consists of the space-time coordinates at which a certain amount of intensity change occurred and the sign (polarity $p_k \in \{+1, -1\}$) of the change.
If the set of events triggered at pixel coordinate $\bfx$ is denoted by $\cE_{\bfx}$, the LTS rendered at an arbitrarily given time, \eg, $t_{\tt{ref}}$, is defined by
\begin{align}
\cT (\bfx, \tref) = \begin{cases}
t_k-\tref, k = \underset{i~\text{for}~\mathbf{e}_i \in \mathcal{E}_{\bfx}}{\text{argmin}}|t_i-\tref| & \text{if}~\cE_{\bfx} \ne \emptyset\\
0 & \text{otherwise}.
\end{cases}
\end{align}
Note that $\bfx$ denotes the raw pixel coordinate, and it can be easily transferred to the normalized coordinate system (\ie~$\bfp$ used across the paper) with intrinsic parameters.

As illustrated in Fig.~\ref{fig:image-like representation}, the proposed LTS is, on one hand, similar to TS in the sense that it also comes with the property of distance field, which enables us to establish event-contour association efficiently. On the other hand, as shown in Fig.~\ref{fig:representation analysis}, the proposed LTS enhances the continuity of the distance field's gradient in a different way. Unlike \cite{zhou2021esvo} that circumvents the unilateral truncation of TS's gradient with a smoothing kernel, LTS simply sets reference time $t_{\tt{ref}}$ as the median timestamp of all involved events.
Such a design brings two benefits:
1) The true location of contour points at reference time is not shifted, and thus, no bias is witnessed in the registration result;
2) The resulting distance transform becomes a signed function that leads to more accurate registration results \cite{zhou2018canny}.

The spatio-temporal registration problem is ultimately formulated as follows.
Let $\cE$ be the involved events\footnote{Involved events are those extracted from the spatio-temporal volume of interest.}.
Assuming the LTS at $t_{\tt{ref}}$ is available, denoted by $\cT_{\tt{ref}}$, the goal is to find the optimal affine model such that all involved events would be warped properly to the zero-value locations of $\cT_{\tt{ref}}$.
The overall objective function of the spatio-temporal registration is
\begin{equation}
\bfa^{\star}= 
\argmin\limits_{\bf{a}} \sum_{\bfe_k \in \cE} \cT_{\tt{ref}}(W (\bfp_k, t_k, t_{\tt{ref}}; \bfa))^2,
\label{eq:energy function of ST registration}
\end{equation}
where $\bfa \doteq \frac{\boldsymbol{\nu}}{Z(t_{\tt{ref}})} = [{a}_x, {a}_y, {a}_z]^\ttT$, and the warping function
\begin{equation}
W(\bfp_k, t_k, t_{\tt{ref}}; \bfa) \doteq \mathcal{A}(\bfp_k; \bfa) (t_{\tt{ref}} - t_k)
\label{eq:warping function}
\end{equation}
transfers events to the common reference time $t_{\tt{ref}}$.
To enhance the smoothness of the object function, we smooth the LTS using a bilateral filter with a Gaussian kernel. 
Besides, an efficient bilinear-interpolation operation is performed to deal with the sub-pixel coordinates in the warping computation of each residual term.
To guarantee the success of Eq.~\ref{eq:energy function of ST registration}, we propose an initialization method that effectively provides an starting point within the convergence basin.

\subsection{Initialization}
\label{subsec:Initialization}
\input{floats/fig_geometric_error}
The motion flow equation (Eq.~\ref{eq:constant parametric model}) provides a possibility to conduct efficient geometric model fitting using optical flow as input measurements.
However, event-based optical flow estimation is an ill-posed problem that is more serve than its standard-vision counterpart.
This is due to the partial observability of event cameras, where events can only be stimulated when the photometric gradient direction is not orthogonal to the epipolar line.
Thus, the recovered motion flow estimate from event data, typically referred to as normal flow, is just the full flow's partial component along the local gradient direction.
Consequently, the linear system (Eq.~\ref{eq:constant parametric model}) established using raw normal flows as alternative measurements does not guarantee a correct fitting result.
Actually, as will be seen in the following analysis, the more the normal flows differ from the full flows, the less accurate the result of model fitting.

To circumvent such a limitation of normal flow, we develop a more effective geometric measurement based on the following fact: The inner product of the full flow vector and the normal flow vector equals to the squared norm of the normal flow vector.
Using as an example the event stimulated at pixel location $\bfp_k$ and by time $t_k$, the full flow vector could be calculated with Eq.~\ref{eq:constant parametric model}, and the above fact is written as
\begin{equation}
\frac{1}{Z_k} {\begin{bmatrix}
-{\nu}_x + x_k {\nu}_z \\ 
-{\nu}_y + y_k {\nu}_z
\end{bmatrix}}^{\ttT} \bfn_k = \bfn_k^{\ttT}\bfn_k,
\label{eq:developed geometric error 1}
\end{equation}
where $\bfn = [n_x, n_y]^{\tt{T}}$ denotes the normal flow.
By simply replacing $\frac{1}{Z(t_k)}$ with $\frac{Z(t_{\tt{ref}})}{Z(t_{\tt{ref}})Z(t_k)}$ in Eq.~\ref{eq:developed geometric error 1} and with some straightforward derivations, we can build the linear system with state vector $\bfa$ as,
\begin{equation}
\begin{bmatrix}
n_{k,x} \\ 
n_{k,y} \\
(\delta t_k \bfn_k - \bfp_k)^{\ttT}\bfn_k 
\end{bmatrix}^{\ttT}\begin{bmatrix}
a_x \\ 
a_y \\
a_z
\end{bmatrix} = -\bfn_k^{\ttT}\bfn_k,
\label{eq:linear system}
\end{equation}
where $\delta t_k \doteq t_{\tt{ref}} - t_k$, and $n_{k,\{\cdot\}}$ denotes the $\{\cdot\}$ component of $\bfn_k$. 
A minimal solver of Eq.~\ref{eq:linear system} requires three measurements, and we use RANSAC~\cite{Fischler81cacm} for robust estimation.

We show the benefit of using the proposed geometric measurement with a toy example. 
As shown in Fig.~\ref{subfig:true and false correspondences}, we simulate a number of contour points (yellow) undergoing a 2D radial expansion.
The true correspondences (green) can be obtained by searching along the full optical flow vectors (blue), while the purple points representing potential correspondences determined by the normal flow vectors (red).
Note that some of the normal flows overlap with the full ones, and such a case can only be witnessed when the local gradient direction is in parallel to the full flow vector.
We compare the results of model fitting using different measurements, including $(i)$ full optical flow, $(ii)$ normal flow, and $(iii)$ the proposed geometric measurement.
As shown in Fig.~\ref{subfig:model fitting analysis of the toy example}, $(i)$ and $(iii)$ return identical model fitting results, while a bias is seen in the result of $(ii)$.
\subsection{Event-Aided Forward Collision Warning System}
\label{subsec:FCW system}
We build an event-aided FCW system on top of the proposed algorithms and a vehicle detection method.
The established FCW system takes as input raw events and intensity images\footnote{Note that intensity images are used only for identifying the leading vehicle.} from either a DAVIS \cite{Lichtsteiner06isscc} sensor or a hybrid camera system made up of a frame-based camera, an event camera, and a beamsplitter~\cite{hidalgo2022event}.
The FCW system manages to output TTC estimates at an ultra framerate.
An overview of the FCW system is given in Fig.~\ref{fig:flowchart}, in which the key modules are highlighted with rectangles in different colors.
Each module takes at least one independent thread and the implementation details are provided in Sec.~\ref{subsec:implementation details}.

%% file: floats/fig_event_representations.tex
\begin{figure}[b]
    \centering
    \begin{subfigure}[t]{0.35\textwidth}
        \centering
        \includegraphics[width=\textwidth]{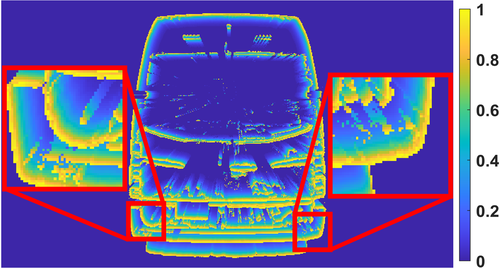}
        \caption{Time Surface \cite{zhou2021esvo, zhou2021emsgc}.}
        \label{fig:Time Surface}
    \end{subfigure}
    \begin{subfigure}[t]{0.35\textwidth}
        \centering
        \includegraphics[width=\textwidth]{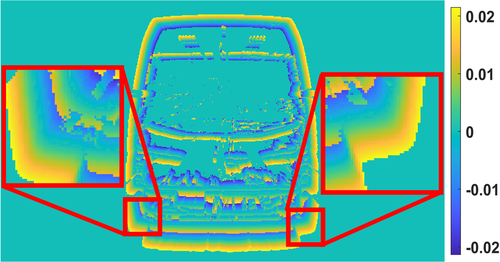}
        \caption{Linear Time Surface.}
        \label{fig:Linear Time Surface}
    \end{subfigure}

\caption{Illustration of two image-like representations of event data. (a) The Time Surface with exponential decay used in \cite{zhou2021esvo, zhou2021emsgc}; (b) The proposed Linear Time Surface.
 The applied color system is explained by the adjacent color bar.
 }
\label{fig:image-like representation}
\end{figure}

%% file: floats/fig_representation_analysis.tex
\begin{figure}[t]
    \centering
    \begin{subfigure}[t]{0.3\textwidth}
        \centering
        \includegraphics[width=\textwidth]{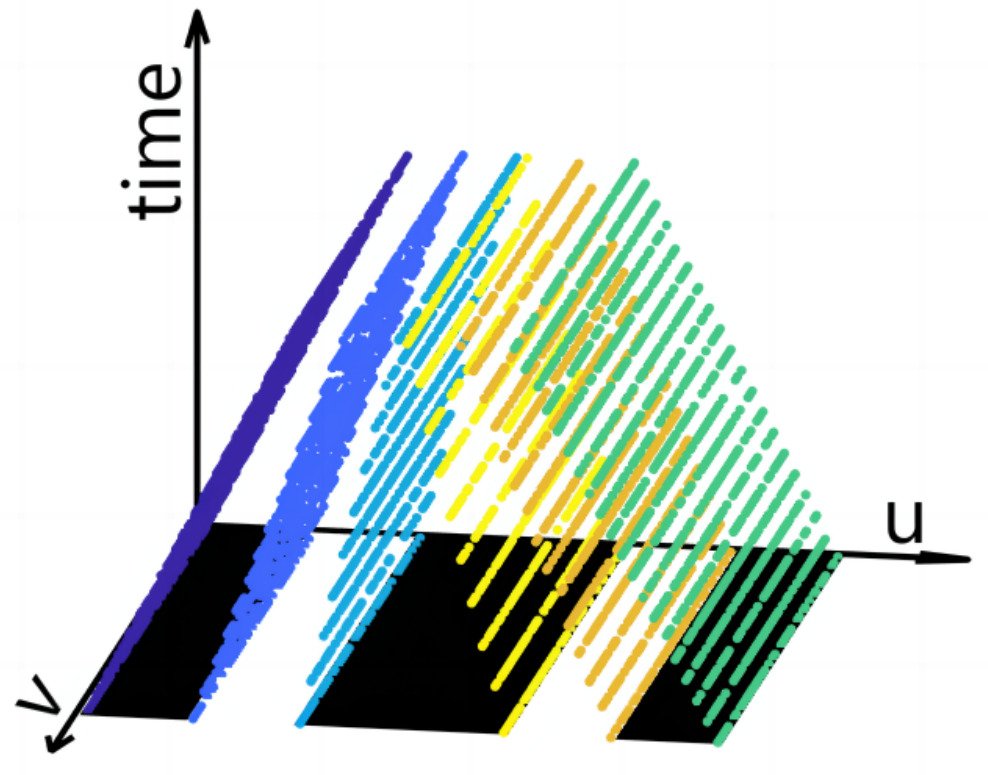}
        \caption{Raw event observations.}
    \end{subfigure}
    \hfill
    \begin{subfigure}[t]{0.3\textwidth}
        \centering
        \includegraphics[width=\textwidth]{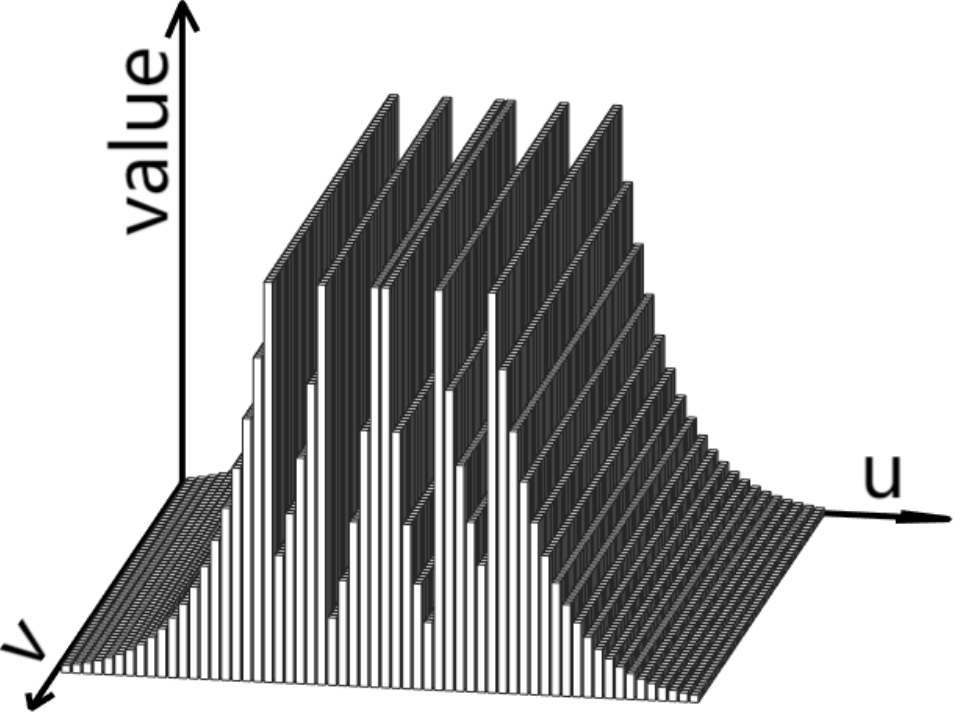}
        \caption{Time Surface.}
    \end{subfigure}
    \hfill
    \begin{subfigure}[t]{0.3\textwidth}
        \centering
        \includegraphics[width=\textwidth]{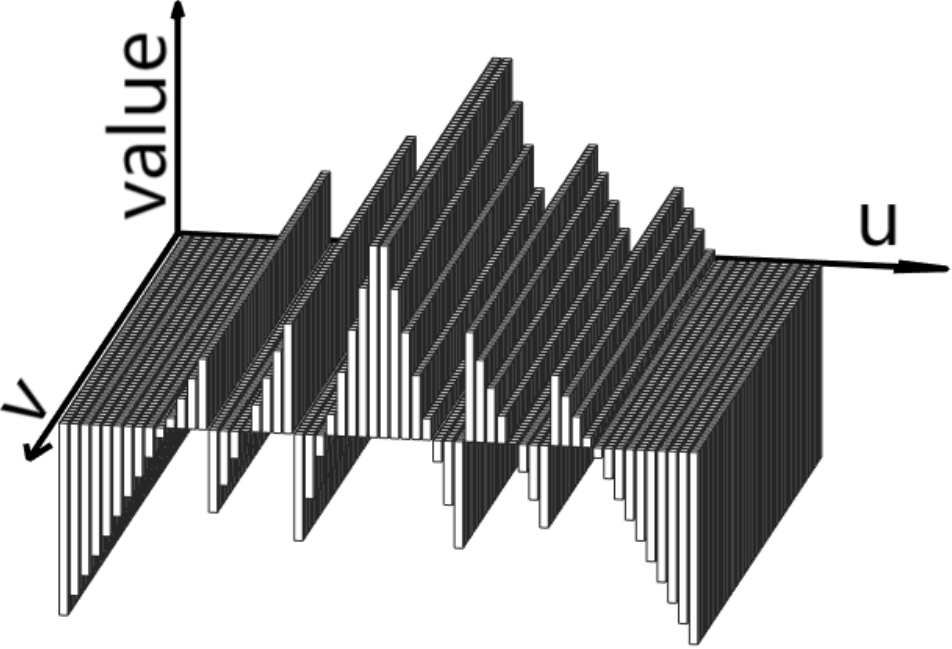}
        \caption{Linear Time Surface.}
    \end{subfigure}
    
    \caption{Comparison on characteristics of TS and LTS.
 (a) Events are triggered in the spatio-temporal domain as six straight lines traversing at different speeds. 
 (b) The generated TS is an unsigned distance transform, and the gradient at the current position of contours is truncated unilaterally.
 (c) The generated LTS is a signed distance transform, and the gradient at the current position of contours is continuous by nature.}
    \label{fig:representation analysis}
\end{figure}

%% file: floats/fig_geometric_error.tex
\begin{figure}[b]
    \centering
    \begin{subfigure}[t]{0.35\linewidth}
        \centering
        \includegraphics[width=\textwidth]{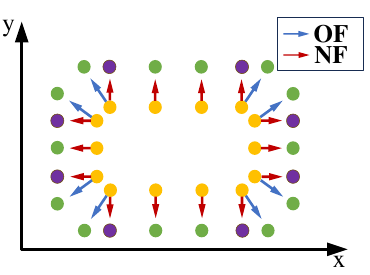}
        \caption{A toy example of radially expanding contour points.}
        \label{subfig:true and false correspondences}
    \end{subfigure}
    \begin{subfigure}[t]{0.35\linewidth}
        \centering
        \includegraphics[width=\textwidth]{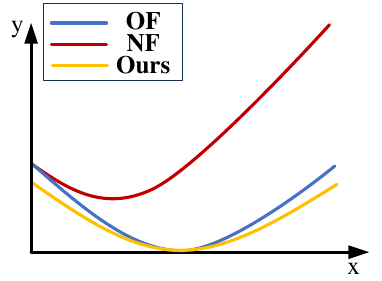}
        \caption{Energy curves of model fitting using different measurements.}
        \label{subfig:model fitting analysis of the toy example}
    \end{subfigure}

\caption{An analysis on performing spatio-temporal registration using different measurements, including $(i)$ optical flow (OF), $(ii)$ normal flow (NF), and $(iii)$ our proposed geometric measurement, respectively.
 Note that only one dimension of the estimated parametric model is visualized in (b) for simplicity.
 }
\label{fig:toy example of registration using different correspondences}
\end{figure}

%% file: chapters/04_experiments.tex
\section{Experiments}
\label{sec:evaluation}

The implementation detail of each core module of the FCW system is first discussed (Sec.~\ref{subsec:implementation details}).
Second, we introduce the datasets used for evaluation (Sec.~\ref{subsec:datasets}).
Finally, both qualitative and quantitative evaluations are performed (Sec.~\ref{subsec:q&q evaluation}), followed with an analysis on the computational performance (Sec.~\ref{subsec:Computational Performance}).

\input{floats/fig_flowchart}
\input{floats/fig_LTS_rendering}

\subsection{Implementation Details}
\label{subsec:implementation details}
\textbf{Identification of Leading Vehicles}.
Although identifying leading vehicles is not the focus of this paper, we do need approaches to detect and keep tracking the leading vehicle in the image plane.
To this end, we utilize YOLO v5 \cite{jocher2022ultralytics} plus DeepSort \cite{Wojke2017simple} that constantly detect and locate the leading vehicle with a bounding box.
The information of bounding box is not only used for indicating the presence of a leading vehicle but also restricting the size of the spatio-temporal volume from which the involved events are extracted. 

\textbf{Rendering of LTS}.
To rationally allocate computational resources while preserving the success of spatio-temporal registration, we adopt an asynchronous strategy for rendering the LTS.
Specifically, the LTS is rendered adaptively according to the expansion rate of the bounding box.
To this end, we propose a simple and effective way to predict the bounding box (i.e., size and location) at anytime by utilizing the latest bounding box and affine model.
Consequently, as an example shown in Fig.~\ref{fig:LTS_render_show}, the frequency of LTS rendering increases adaptively when the two vehicles get closer with no deceleration.

\textbf{Robust Sampling.}
Although most of the background events are excluded by the bounding box, we still need a sampling strategy that further resists noises and outliers.
To this end, we calculate the first-order and second-order image gradients of the LTS.
Events will be reserved on the condition that the first-order gradient's magnitude is larger than $1 \times 10^{-5}$, and the second-order one's magnitude is lower than $1 \times 10^{-3}$. 

\textbf{Hyperparameters}.
We implement our robust linear solver on top of a standard RANSAC framework.
Specifically, the maximum number of iterations is set to 300, and the inlier ratio is 0.9.
As for the nonlinear spatio-temporal registration, we apply the Levenberg-Marquardt (LM) algorithm to iteratively solve Eq.~\ref{eq:energy function of ST registration}.
Once initialized, 10 iterations are typically enough for convergence.

\subsection{Datasets}
\label{subsec:datasets}
Three datasets are used to evaluate our method, and all these datasets capture scenarios featuring a leading vehicle followed by the host vehicle driving in the same lane.
The first one is a fully synthetic dataset created using CARLA \cite{dosovitskiy2017carla}, an open-source simulator for urban and suburban driving (see Fig.~\ref{fig:carla_show}).
We synthesize a hetero system that consists of a standard RGB camera and an event-based camera, which share identical intrinsic and extrinsic parameters (w.r.t the body coordinate of the host vehicle).
The synthetic dataset consists of three subsets featuring different motion patterns, scenes, and different types of leading vehicles (\eg~Sedan, SUV, and Van, etc.). 
The second dataset is obtained by using a small-scale test platform that simulates the configuration of the TTC task (see Fig.~\ref{fig:sensor of slider rail}).
The platform contains a hybrid optic system and a motorized slider.
The hybrid optical system, mounted on the slider, consists of a standard camera, an event camera, and a beamsplitter.
By spliting the incoming ray into two directions, the beamsplitter ensures both cameras share the same field of view.
The position and the velocity of the slider can be controlled to emulate the relative motion between the leading vehicle (a model car) and the observer, and thus, the groundtruth TTC can be easily obtained.
The third dataset, denoted by FCWD, consists of real data collected using a multi-sensor suite (including a pair of standard cameras, a pair of event cameras, and a LiDAR) as shown in Fig.~\ref{fig:sensor of fcwd}.
The multi-sensor suite is mounted on the host vehicle that approaches the leading one at different speed.
The groundtruth TTC can be obtained by differentiating the LiDAR data between the two cars.
The characteristics of each dataset can be found in Table.~\ref{tab:dataset characteristics}, and are detailed in the supplementary material.

\input{floats/fig_dataste_sensor_setup}
\input{floats/table_dataset_characteristics}

\subsection{Quantitative and Qualitative Evaluation}
\label{subsec:q&q evaluation}
To prove the effectiveness of the proposed geometric measurement, we first compare the performance of our initialization method using as measurements the normal flow (NF), the optical flow (OF), and the proposed geometric measurement (Ours), respectively.
Note that the normal flows are estimated from event data using the method in~\cite{benosman2013event}, and the optical flows are approximated by using feature correspondences on intensity images\footnote{Note that intensity images are used only for providing full optical flow in the comparison, and they are not available in our method for geometric model fitting.}.
We run the three initializers on all sequences of the synthetic data and analyze the statistics of the results.
For quantitative evaluation, we use as metric the relative TTC error, defined as $e_{\text{TTC}} = \vert \frac{t_{gt} - t_{est}}{t_{gt}} \vert \times 100\%$,
where $t_{gt}$ denotes the GT TTC value and $t_{est}$ the estimated TTC value.
As shown in Fig.~\ref{fig:Evaluation of the initialization method with different measurements}, our initializer is on par with the OF-based one, and both outperform remarkably the NF-based one.
Besides, we compare the resulting geometric models by illustrating the spatially-and-temporally aligned events.
As seen in Fig.~\ref{fig:event alignment comparison}, the model fitted using our geometric measurement leads to better alignment of events and a higher contrast.
\input{floats/fig_carla_initializer_experiment}

We also compare our method against several alternative solutions in terms of TTC accuracy and computational efficiency.
Methods used for comparison consist of the \textit{CMax} method \cite{Gallego18cvpr}, an improved version of \textit{CMax} initialized using our initializer, an event-based method (\textit{FAITH}~\cite{dinaux2021faith}) based on Focus of Expansion (FoE) estimation, an image-based method (\textit{Image's FoE}~\cite{stabinger2016monocular}) that also uses FoE estimation, and an open-source event-based method called \textit{ETTCM}~\cite{nunesTimeToContact2023}.

\input{floats/table_counterpart_comparison_simdata}
\input{floats/table_counterpart_comparison_realdata}

We report the average value of $e_{\text{TTC}}$ and runtime of each method.
As shown in Table.~\ref{tab:sim_quantitative_evaluation} and Table.~\ref{tab:real_quantitative_evaluation}, \textit{CMax} and its initialized version generally achieve the best accuracy.
This is because \textit{CMax} uses all events within the spatio-temporal volume defined by the bounding box, enhancing accuracy with a better signal-to-noise ratio.
However, the non-least-squares nature of \textit{CMax} results in high computation complexity, making it unsuitable for real-time applications like TTC estimation (\eg, it takes \textit{CMax} 3 s to return an result from $2e5$ events occurred during an interval of 18 ms).
\textit{FAITH} shows a potential for real-time applications.
However, its success heavily relies on the sense that the normal flow distributed uniformly around the FoE.
Additionally, simply using normal flow as a replicate of optical flow leads to inaccurate TTC estimation results.
The \textit{Image's FoE} method performs normally good in terms of both metrics.
However, it requires two successive images as input, and the runtime reported does not consider the time interval between two successive exposures ($30\sim100$ ms).
This system latency induced by standard cameras can be a fatal delay to online control and decision making.
As an event-based incremental methodology, \textit{ETTCM} does not scale well with input data generated per unit time.
Our method outperforms \textit{ETTCM} in both metrics.
Thanks to the robust sampling strategy and efficient initializer, our method achieves stable runtime in various scenarios.
Since the input of each method is different, it is hard to set a criterion for counting runtime in an absolutely fair manner.
Most methods for comparison are run with default parameters.
Some of the event-based methods (\eg, \textit{CMax}) need a fine tune on the number of events used according to the spatial resolution of the event camera used.
The mean runtime is reported for each sequence, and detailed configuration of each method can be found in the supplementary material.
In conclusion, our method balances accuracy and efficiency.

In addition, we also conduct an experiment to investigate how sensitive our method is to the lateral offset of the preceding vehicle.
We add a range of lateral offsets (\ie, $[-4, 4]$ \text{m}) to the preceding vehicle in our synthetic dataset, and evaluate the performance of our TTC method.
As seen in Fig.~\ref{fig:lateral_offset_evaluation}, the relative TTC error is always below $10\%$ as long as the lateral offset is smaller than 3 m, indicating that our method has certain robustness against violation of the assumption on application scenarios.

\input{floats/figtable_lateral_offset_computational_performance}

\subsection{Computational Performance}
\label{subsec:Computational Performance}
The proposed FCW system is implemented in C++ on ROS and runs in real-time on a laptop with an AMD Ryzen 5800H CPU and an NVIDIA RTX 3060 GPU. 
GPU is only used for Vehicles Identification node.
The computational performance of each node is summarized in Table.~\ref{tab:demo_computational_efficiency}.
Note that these numbers reported are obtained when using an RGB camera ($1440 \times 1080$ pixels) and an event camera ($640 \times 480$ pixels).
Every node takes one independent thread, and their runtime are listed in the right-most column.
As observed in our experiments, the initialization is usually applied only once, and all subsequent updates can be done by constantly calling the spatio-temporal registration.
Thus, our system can output TTC up to 200 Hz.

It is also worth mentioning that the advantage of our method is not only in terms of computation efficiency, but also in terms of system latency. 
If the exposure time of a standard camera is ignored, the latency of pure image-based methods consists of the time interval between two successive exposures (e.g., 33 ms for a 30-fps camera), and the computation time (e.g., 40 ms by \cite{stabinger2016monocular,poiesi2016detection}). 
The overall system latency is 73 ms. 
For our method, the average latency of our algorithm is mainly caused by the computation time (5-10 ms once initialized), indicating a great potential for handling sudden variations in relative speed.

%% file: floats/fig_flowchart.tex
\begin{figure}[t!]
  \centering
  \includegraphics[width=0.6\linewidth]{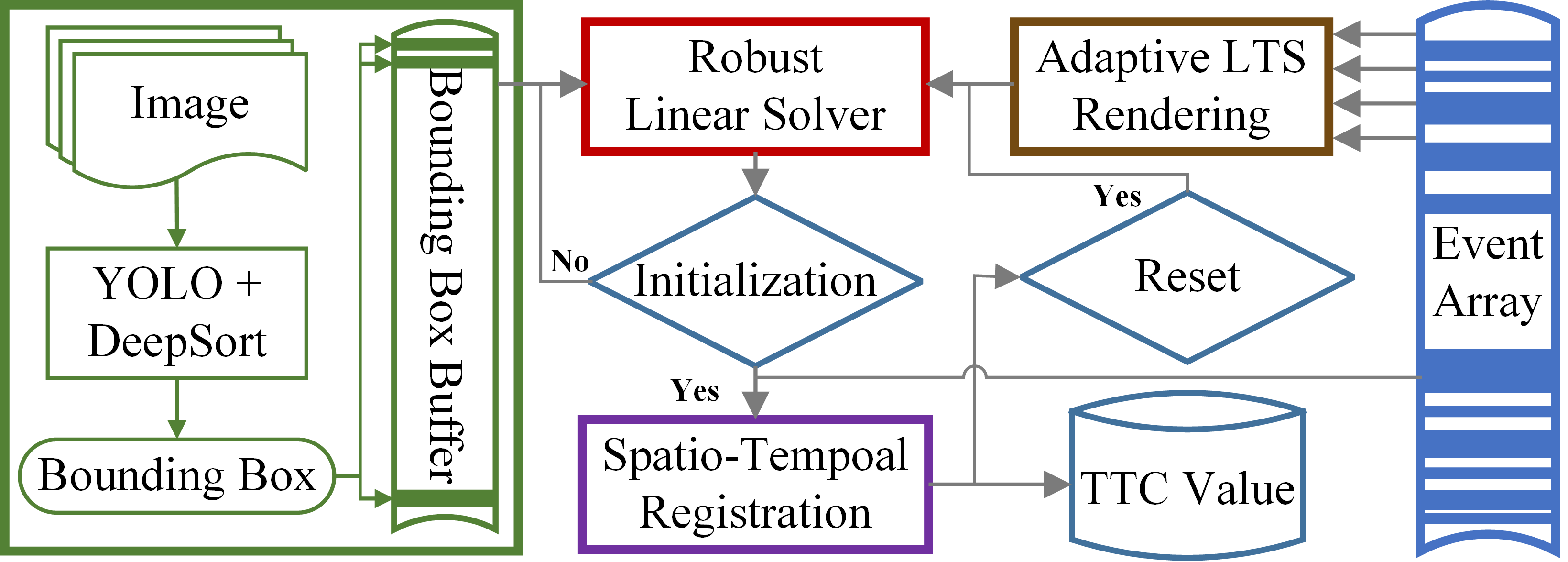}
  \caption{Flowchart of the proposed FCW system.
  Key modules of the system consist of the vehicle detection module (green), the LTS rendering module (brown), the robust linear solver (red), and the spatio-temporal registration (purple).}
  \label{fig:flowchart}
\end{figure}

%% file: floats/fig_LTS_rendering.tex
\begin{figure}[t!]
    \centering
    \includegraphics[width=0.8\textwidth]{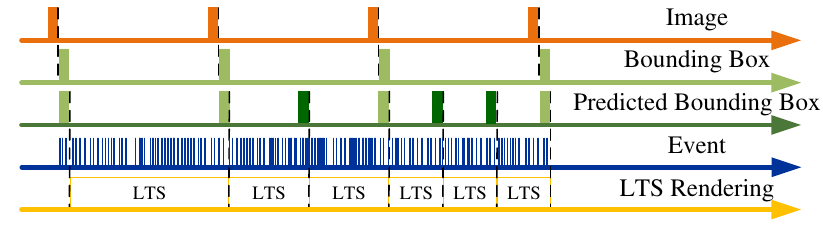} 
    \captionsetup{font=small} 
    \caption{Illustration of the adaptive LTS rendering.}
    \label{fig:LTS_render_show}
\end{figure}

%% file: floats/fig_dataste_sensor_setup.tex
\begin{figure}[t]
    \centering
    \begin{subfigure}[t]{0.25\textwidth}
        \frame{\includegraphics[width=\textwidth]{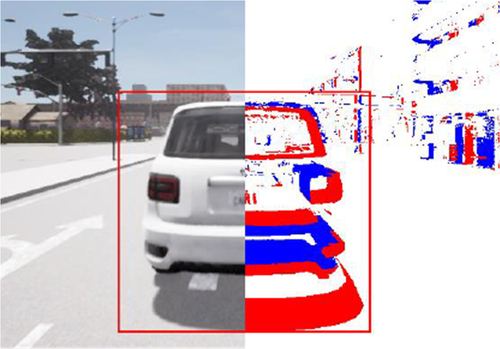}}
        \caption{Synthetic dataset.}
        \label{fig:carla_show}
    \end{subfigure}
    \begin{subfigure}[t]{0.385\textwidth}
        \frame{\includegraphics[width=\textwidth]{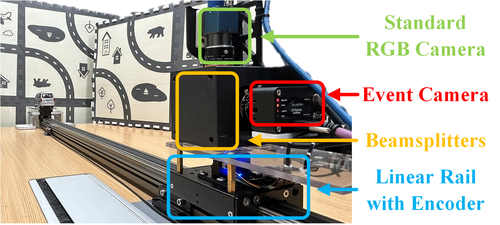}}
        \caption{Small-scale test platform.}
        \label{fig:sensor of slider rail}
    \end{subfigure}
    \begin{subfigure}[t]{0.29\textwidth}
        \frame{\includegraphics[width=\textwidth]{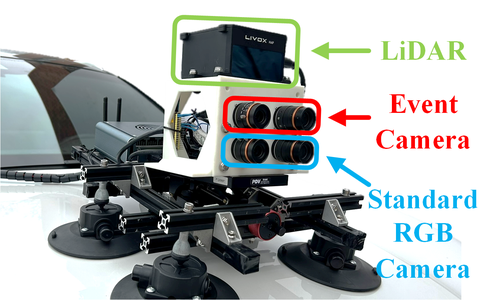}}
        \caption{Multi-sensor suite.}
        \label{fig:sensor of fcwd}
    \end{subfigure}
    \caption{
    Illustration of our datasets and devices for data collection.
    (a) A selected snapshot of the synthetic dataset, on each side of which the intensity information and the events (with a naive accumulation) are illustrated, respectively.
    (b) The configuration of the small-scale test platform.
    (c) The multi-sensor suite mounted on a car.}
    \label{fig:real dataset sensor setup}
\end{figure}

%% file: floats/table_dataset_characteristics.tex
\begin{table}[b]
\centering
\caption{Characteristics of our datasets.} 
\begin{adjustbox}{max width=\linewidth}
\setlength{\tabcolsep}{3pt}
\begin{tabular}{c|cccccc}
\toprule
Dataset Name   & 
Platform&
Terrain   &
\begin{tabular}[c]{@{}c@{}} Sensor \\ Motion \end{tabular} &
\begin{tabular}[c]{@{}c@{}} Event \\ Cameras\end{tabular} &
\begin{tabular}[c]{@{}c@{}} RGB \\ Cameras\end{tabular} &
\begin{tabular}[c]{@{}c@{}} Total \\ Sequences\end{tabular} \\
\midrule
Synthetic & \multicolumn{1}{c}{CARLA} & 
\begin{tabular}[c]{@{}c@{}} Urban \\ Suburban \end{tabular} &
\begin{tabular}[c]{@{}c@{}} Const. Vel \\ Acceleration \end{tabular} &
\begin{tabular}[c]{@{}c@{}} CARLA DVS \\ $640\times480$ \end{tabular} &
\begin{tabular}[c]{@{}c@{}} CARLA RGB \\ $640\times480$ \end{tabular} &
21 \\
\midrule
Slider & \multicolumn{1}{c}{Rail \& Slider} & 
Indoor &
Const. Vel &
\begin{tabular}[c]{@{}c@{}} inivation DVXplorer \\ $640\times480$ \end{tabular} &
\begin{tabular}[c]{@{}c@{}} DAHENG MER2 \\ $1440\times1080$ \end{tabular} &
3 \\
\midrule
FCWD & \multicolumn{1}{c}{Car} & 
Urban &
Normal Driving &
\begin{tabular}[c]{@{}c@{}} Prophesee EVKv4 $\times2$\\ $1280\times720$ \end{tabular} &
\begin{tabular}[c]{@{}c@{}} FLIR Blackfly S $\times2$ \\ $1920\times1200$ \end{tabular} &
3 \\

\bottomrule 
\end{tabular}
\end{adjustbox}
\label{tab:dataset characteristics}
\end{table}

%% file: floats/fig_carla_initializer_experiment.tex
\begin{figure}[t!]
    \centering
    \begin{subfigure}[t]{0.45\linewidth}
        \includegraphics[width=\textwidth]{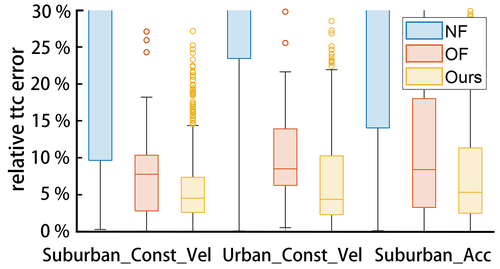}
        \caption{Statistics of TTC results.
        }
        \label{fig:Evaluation of the initialization method with different measurements}
    \end{subfigure}
    \begin{subfigure}[t]{0.4\linewidth}
        \frame{\includegraphics[width=\textwidth]
        {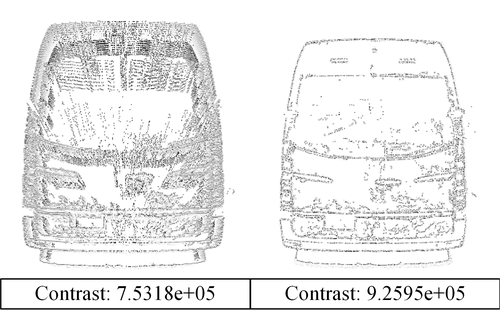}}
        \caption{Event alignment result.}
        \label{fig:event alignment comparison}
    \end{subfigure}
    \caption{Comparison of our initialization method using different measurements. 
    (a): Statistics of the relative TTC error from different methods illustrated with a box plot.
    The boxes and whiskers of NF's results are not shown completely for compactness. 
    (b): Results of event alignment (in the format of images of warped events) with models fitted using NF (left) and the proposed geometric measurement (right), respectively.}
\end{figure}

%% file: floats/table_counterpart_comparison_simdata.tex
\renewcommand{\arraystretch}{1.2}
\begin{table}[t]
\scriptsize
\caption{Quantitative evaluation on synthetic data.}
\label{tab:sim_quantitative_evaluation}
\centering
\setlength{\tabcolsep}{0.01\linewidth}
\begin{tabular}{l|cc|cc|cc}
\Xhline{1pt}
\multirow{2}{*}{\textbf{Method}}
&\multicolumn{2}{c|}{\emph{Suburban\_Const\_Vel}}
&\multicolumn{2}{c|}{\emph{Urban\_Const\_Vel}}
&\multicolumn{2}{c}{\emph{Suburban\_Acc}}\\
\cline{2-7}
&$e_{\text{TTC}}$     &Runtime (s)
&$e_{\text{TTC}}$     &Runtime (s)
&$e_{\text{TTC}}$     &Runtime (s)\\
\Xhline{0.7pt}

CMax\cite{Gallego18cvpr}
& 3.76\%    & 1.37
& 6.80\%    & 0.94
& 5.06\%    & 1.32 \\
Our Init + CMax~\cite{Gallego18cvpr}
& 3.25\%    & 0.82
& 5.87\%    & 0.54
& 4.03\%    & 0.68\\
FAITH~\cite{dinaux2021faith}
& 88.6\%   & 0.208
& 60.3\%   & 0.176
& 87.2\%   & 0.155  \\
Image's FoE~\cite{stabinger2016monocular}
& 7.80\%    & 0.037  
& 8.46\%    & 0.039       
& 8.59\%    & 0.037 \\
ETTCM~\cite{nunesTimeToContact2023}
&  7.88\%   &   0.031
&  8.43\%   &   0.034 
&  7.90\%   &   0.029 \\
Ours
& 4.29\%    &   0.009
& 4.78\%    &   0.008
& 3.58\%    &   0.01 \\

\Xhline{1pt}
\end{tabular}
\end{table}

%% file: floats/table_counterpart_comparison_realdata.tex
\renewcommand{\arraystretch}{1.2}
\begin{table}[t]
\scriptsize
\caption{Quantitative evaluation on real data.}
\label{tab:real_quantitative_evaluation}
\centering
\setlength{\tabcolsep}{0.01\linewidth}
\begin{tabular}{l|cc|cc|cc}
\Xhline{1pt}
\multirow{2}{*}{\textbf{Method}}
&\multicolumn{2}{c|}{\emph{Slider\_500}}
&\multicolumn{2}{c|}{\emph{Slider\_750}}
&\multicolumn{2}{c}{\emph{Slider\_1000}}\\
\cline{2-7}
&$e_{\text{TTC}}$     &Runtime (s)
&$e_{\text{TTC}}$     &Runtime (s)
&$e_{\text{TTC}}$     &Runtime (s)\\
\Xhline{0.7pt}
CMax\cite{Gallego18cvpr}
&  4.18\%   & 0.90
&  4.16\%   & 0.85 
&  2.74\%   & 0.93 \\
Our Init + CMax~\cite{Gallego18cvpr}
&  4.18\%    & 0.63 
&  4.15\%    & 0.76 
&  2.74\%    & 0.57 \\
FAITH~\cite{dinaux2021faith}
& 37.58\%   &  0.151
& 34.45\%   &  0.186
& 47.94\%   &  0.233\\
Image's FoE~\cite{stabinger2016monocular}
& 12.37\%      & 0.013 
& 11.35\%      & 0.013 
&  7.55\%      & 0.012 \\
ETTCM~\cite{nunesTimeToContact2023}
& 16.12\%           & 0.405
& 18.99\%           & 0.498
& 15.20\%           & 0.191 \\
Ours
&  10.02\%    & 0.017
&   8.95\%    & 0.015 
&  12.43\%    & 0.016  \\

\Xhline{1pt}

\multirow{2}{*}{\textbf{Method}}
&\multicolumn{2}{c|}{\emph{FCWD 1}}
&\multicolumn{2}{c|}{\emph{FCWD 2}}
&\multicolumn{2}{c}{\emph{FCWD 3}} \\
\cline{2-7}
&$e_{\text{TTC}}$     &Runtime (s)
&$e_{\text{TTC}}$     &Runtime (s)
&$e_{\text{TTC}}$     &Runtime (s)\\
\Xhline{0.7pt}
CMax\cite{Gallego18cvpr}
&   2.42\%  &  2.53
&   2.39\%  &  3.12
&   3.04\%  &  3.34\\
Our Init + CMax~\cite{Gallego18cvpr}
&  2.33\%  &  1.94
&  2.26\%  &  2.38
&  2.97\%  &  2.59\\
FAITH~\cite{dinaux2021faith}
& 25.49\%    & 0.214 
& 27.91\%    & 0.203 
& 39.15\%    & 0.210 \\
Image's FoE~\cite{stabinger2016monocular}
& 5.39\%   & 0.017 
& 5.70\%   & 0.018
& 6.02\%   & 0.017 \\
ETTCM~\cite{nunesTimeToContact2023}
&   15.52\%         & 0.307
&   18.45\%         & 0.282
&   19.08\%         & 0.295 \\
Ours
&    9.84\%    & 0.017 
&   11.55\%    & 0.023 
&   14.09\%    & 0.025 \\

\Xhline{1pt}
\end{tabular}
\end{table}

%% file: floats/figtable_lateral_offset_computational_performance.tex
\begin{figure}[t]
    \centering
    \begin{minipage}{.45\linewidth}
        \centering
        \includegraphics[width=5.5cm]{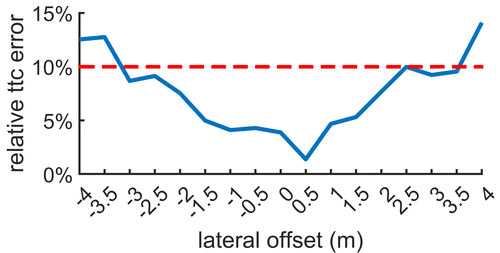}
        \caption{Sensitivity analysis on the lateral offset of the preceding car.}
        \label{fig:lateral_offset_evaluation}
    \end{minipage}
    \begin{minipage}{.45\linewidth}
        \centering
        \resizebox{.9\textwidth}{!}{
        \begin{adjustbox}{max width=\linewidth}
        \begin{tabular}{ccc}
        \Xhline{1.5pt}
        \multicolumn{1}{c}{\textbf{Node ( \#Threads)}} &
        \multicolumn{1}{c}{\textbf{Function}} & \multicolumn{1}{c}{\textbf{Time (ms)}} \\ \Xhline{1.5pt}
        \begin{tabular}[c]{@{}c@{}} \textbf{Vehicles} \\ \textbf{Identification (GPU)}\end{tabular}
        & Yolov5  & 11\\ 
        & DeepSort  & 6\\ 
        \Xhline{0.7pt}
        \textbf{Data Pre-processing (1)} & LTS Rendering  & 1 - 3 \\
        \Xhline{0.7pt}
        \begin{tabular}[c]{@{}c@{}} \textbf{Initialization} \\ \textbf{\& Reset (1)}\end{tabular} & Robust Sampling 
        & 5\\
         & Linear Solver   & 10\\
         \Xhline{0.7pt}
         \begin{tabular}[c]{@{}c@{}} \textbf{Spatio-Temporal} \\ \textbf{Registration (1)}\end{tabular}
        &   Non-Linear Solver & 5 - 10\\ 
        \Xhline{1.5pt}
        \end{tabular}
        \end{adjustbox}
        }
        \captionof{table}{Computational performance.}
        \label{tab:demo_computational_efficiency}
    \end{minipage}
\end{figure}

%% file: chapters/05_conclusions.tex
\section{Conclusion}
\label{sec:conclusion}
This paper proposes an event-based method for estimating the time to a potential collision with a preceding vehicle.
To the best of our knowledge, this is the first event-based TTC solution for autonomous/assisted driving scenarios.
The proposed two-step approach can estimate efficiently the TTC via fitting a geometric model to raw event data.
Experiments demonstrate the effectiveness of the proposed method on both synthetic and real data, showing that our method has the best overall performance in terms of accuracy and speed.
We will release the software together with the datasets to the community and wish that the work will serve as a benchmark for future research on event-based TTC estimation.

%% file: chapters/05a_acknowledgment.tex
\section*{Acknowledgment}
We thank Javier Hidalgo-Carri{\'o} and Davide Scaramuzza for releasing the design of the Beamsplitter~\cite{hidalgo2022event}, based on which we build our FCW system. 
We also thank Dr. Yi Yu for proof reading.
This work was supported by the National Key Research and Development Project of China under Grant 2023YFB4706600.

%% file: chapters/06_appendix.tex
\section*{Overview}
\label{sec:overview}

We provide more details about the submission in this document, which includes:
\begin{itemize}
\item A justification of the proposed flow-dynamics consistent geometric model. (Sec.~\ref{sec:key_iinsights})
\item An in-depth look at the three datasets we generate/collect. (Sec.~\ref{sec:dataset})
\item More detailed results on real-world datasets. (Sec.~\ref{sec:real_world_experiments})
\item An additional discussion on the performance of all involved methods for comparison, explaining the number reported in the paper according to an ablation study. (Sec.~\ref{sec:quantitative_experiment})
\end{itemize}

%+++++++++++++++++++++++++++++++++++++++++++++++++++++++++++++++++++++++++

\section{Justification of the Flow-Dynamics Consistent Model}
\label{sec:key_iinsights}

As shown in~\cite{Gallego18cvpr}, the success of geometric-model fitting to event data hinges on using an accurate parametric model.
To this end, we develop a time-variant affine model that captures the true flow-field dynamics.
To justify, we add a comparison of our proposed time-variant affine model against the widely used constant affine model, and also, the simplified affine model (assuming no horizontal motion) by \textit{ECMD}~\cite{mcleod2023globally}.
The evaluation metric used is the contrast of the resulting image of warped events (IWE), and the goodness of fitting can also be assessed qualitatively from the IWE's sharpness.
As seen in Fig.~\ref{fig:warping_model}, our method outperforms the others, justifying the key insight of our method.

\input{floats/fig_warping_model_supp}

%+++++++++++++++++++++++++++++++++++++++++++++++++++++++++++++++++++++++++

\section{Dataset}
\label{sec:dataset}

\input{floats/fig_data_seq}

We elaborate on the three datasets introduced in Section 4.2 of the paper. 
Our investigation reveals a significant lack of specific event datasets for the task of time-to-collision estimation. 
To this end, we create three platforms (1 virtual + 2 real) for data generation, and they consist of: 1) A customized virtual environment that synthesizes data in traffic scenes (\ref{subsec:synthetic dataset});
2) A small-scale test platform that mimics the discussed scenarios (\ref{subsec:slider}); and 3) A multi-sensor suite employed on a real car for data collection (\ref{subsec:fcwd}).
Fig.~\ref{fig_data_seq} provides a snapshot of the platform and generated data for each dataset, and Tab.~\ref{tab:hardware_specification} details the configuration of the sensors used.
We detail the data generation process for each dataset in the following.
\input{floats/table_hardware_specifications}

\subsection{Synthetic Dataset}
\label{subsec:synthetic dataset}

The simulation of forward-collision scenarios is built on top of CARLA\cite{dosovitskiy2017carla}, an open-source simulation platform. 
CARLA offers extensive APIs for customizing vehicle motions and scenes, supporting a wide array of sensors commonly used in robotics, including RGB cameras, LIDAR, IMU, and also, event-based cameras. 
Three subsets are created, and each one features distinct motion patterns and scenes. 
We set the synthesized event camera's parameters as follows.
The $\mathrm{eps}$ value is set to $0.3$, the refractory period is $1 \times 10^{-5}$ seconds, and the contrast threshold is $0.15$. 
The calculation of the groundtruth TTC is based on the absolute distance and relative speed between the host vehicle and the preceding one. 

\subsection{Slider Dataset}
\label{subsec:slider}
\input{floats/fig_slider_demonstrate}
To narrow the gap between simulated data and real-world ones, we design a small-scale test platform using a miniature replica of real vehicles to simulate car crash scenarios.
As shown in Fig.~\ref{fig:slider_demonstrate}, this test platform is composed of a motorized slider, a hybrid optical system based on an open-source design in~\cite{hidalgo2022event}, and a 1:24 scale vehicle model. 
The hybrid optical system consists of an inivation DVXplorer event camera of $640\times480$ pixel resolution, an RGB camera with a spatial resolution of $1440\times1080$ pixel, and a beamsplitter that divides incoming light into two paths, ensuring a unified field of view for both cameras.
For precisely identifying event points within the bounding box of the leading vehicle, it is important to establish a pixel-to-pixel correspondence map, and furthermore, synchronize the time clocks of the two heterogeneous sensors.
This pixel-to-pixel mapping between the two cameras can be established through an offline calibration scheme. 
To temporally synchronize these cameras, we use an STM32 development board, which sends a 25-Hz clock signal that triggers both cameras, ensuring a precise synchronization. 
The hybrid optical system, mounted on a slider, simulates collisions at three different speeds (\ie, 500 mm/s, 750 mm/s, and 1000 mm/s, respectively). 
The groundtruth TTC is derived from the slider's position along the rail and its speed measured by the motor encoder.

\subsection{Forward Collision Warning Dataset (FCWD)}
\label{subsec:fcwd}

\input{floats/table_dataset_comparision}

Publicly available datasets for autonomous driving using event cameras fall short of addressing the specific requirements of TTC estimation research. 
As listed in Table~\ref{tab:dataset comparision}, these datasets either exhibit low spatial resolution (\eg, $346\times260$ pixel or $640\times480$ pixel) or lack comprehensive temporal synchronization among sensors.
Additionally, these datasets are predominantly designed for tasks such as Simultaneous Localization and Mapping (SLAM) and object detection, rather than for TTC estimation. 
To achieve a wider Horizontal Field of View (HFOV), all datasets utilize short focal length lenses for both event cameras and RGB cameras. 
Moreover, the host vehicle, on which these sensors are mounted, typically decelerates in advance to keep a safe distance from the leading vehicle.

To this end, we develop a multi-sensor suite incorporating a stereo event camera, a stereo RGB camera, and a LiDAR. 
Specifically, we employ a HD-resolution (720p) event camera Prophesee EVKv4 to record event data.
The event camera is synchronized with other sensors to millisecond accuracy via a triggering signal from an STM-32 development board. 
The RGB camera and the LiDAR are synchronized to sub-millisecond accuracy using the Precision Time Protocol (PTP). 
Additionally, we equip event cameras and RGB cameras with telephoto lenses, suitable for data collection in forward-collision scenarios.
Note that we only use one event camera and one RGB camera in this work.

Three sequences were recorded using our multi-sensor suite to evaluate our algorithm in real-world settings. 
The multi-sensor system is mounted on the host vehicle's engine cover with suction cups. 
The vehicle is driven towards a stationary one ahead, breaking only at the minimum safety distance, as depicted in the multimedia content.
Groundtruth TTC values are obtained from the distance to the leading vehicle, measured by the LiDAR, and the vehicle’s speed, determined by the LiDAR-inertial odometry (Fast-lio~\cite{xu2021fast}). 

%+++++++++++++++++++++++++++++++++++++++++++++++++++++++++++++++++++++++++

\section{Detailed Real-World Experiments of TTC Estimation}
\label{sec:real_world_experiments}

We provide more detailed results on our real data (\textit{Slider} and \textit{FCWD}).
As shown in Fig.~\ref{fig_strttc_experiment_result}, our results are always consistent with the ground truth. 
We observe that the TTC results become increasingly accurate as the distance between the host vehicle and the leading vehicle decreases.
This happens due to the fact that the contour of the leading vehicle on the image plane enlarges, and it will generate more event data, leading to linear time surfaces (LTS) with more spatio-temporal information.
We contend that improvements are feasible with further engineering efforts.
Given the current runtime statistics for a single computation, an updating rate of 200 Hz, as claimed in the paper, can be achieved.

\input{floats/fig_strttc_experiment_result}

%+++++++++++++++++++++++++++++++++++++++++++++++++++++++++++++++++++++++++

\section{Discussion on Accuracy and Efficiency}
\label{sec:quantitative_experiment}

This section elaborates on detailed configuration of each method we compare against in Section 4.3 of the paper.
The configuration and parameter selection for each method significantly affect the accuracy of TTC calculations and runtime.
For a fair comparison, Table.~\ref{tab:algorithm_comparison} shows the result of each method under various parameter settings on our FCWD dataset.
The runtime reported in the table represents the average computation time to get one TTC result on the corresponding sequence.

Image’s FoE~\cite{stabinger2016monocular} takes as input the intensity images from an RGB camera running at 10 Hz.
Within the bounding box, the SURF~\cite{Bay08cviu} algorithm is employed to extract feature points on the lead vehicle in two consecutive frames. 
Based on the matched feature points, the affine motion model is estimated to calculate the TTC.
The runtime includes feature extraction and matching between two consecutive frames, as well as the affine model fitting and the TTC calculation.
Note that the runtime does not include the system latency caused by the time interval between two successive exposures.

\input{floats/table_algorithm_comparison}
\input{floats/fig_runtime_accuracy_comparison}

In FAITH~\cite{dinaux2021faith}, we employ default parameters from its open-source code.
Events triggered within the bounding box are used as input, and the runtime represents the average time for each result.

For the methods of CMax~\cite{Gallego18cvpr} and Our Init + CMax, the main factor affecting the accuracy of TTC estimation and computation time is the number of events involved.
There are two main strategies: using a fixed temporal window or a constant number of events.
In the forward-collision scenario, the number of events within the bounding box fluctuates significantly over time with a fixed time interval on our high-resolution event camera.
At the beginning of each sequence, the camera is far from the leading vehicle and only a small number of events are triggered, resulting in large estimation error. 
As the leading vehicle gets closer, an increasingly larger number of events are generated in a short period of time, resulting in long computation time or even an abortion of the algorithm.
Therefore, we choose the second strategy, \ie, using a constant event number of events.
Table.~\ref{tab:algorithm_comparison} shows the estimation error under different event numbers. 
To seek the balance between the computation time and estimation accuracy, we report the result of $2e^5$ in the experiment result.

The ETTCM~\cite{nunesTimeToContact2023} method supports different motion models and neighboring sizes.
We follow evaluate its performance under different configurations.
Our evaluation tries three motion models with a neighboring sizes of 2 and 3, respectively, looking for a comprehensive performance in terms of accuracy and efficiency. 
The combination of a scaling model and a neighbouring size of 2 is selected in the report of our paper.
The ETTCM method estimates the TTC and reports computation time on a per-event basis. 
To simplify comparisons, we define the computation time of the ETTCM method as the multiplication of the time required by a single computation and the total number of events processed by our method every time. 
This offers a standardized metric for comparing the runtime of our approach against that of the ETTCM method.

Figure~\ref{fig:runtime_accuracy} presents a comparison of runtime and accuracy for all event-based TTC estimation methods, indicating our method achieving a state-of-the-art performance.

%+++++++++++++++++++++++++++++++++++++++++++++++++++++++++++++++++++++++++

%% file: floats/fig_warping_model_supp.tex
\begin{figure}[b]
    \small
    \begin{center}
        \includegraphics[width=0.90\linewidth]{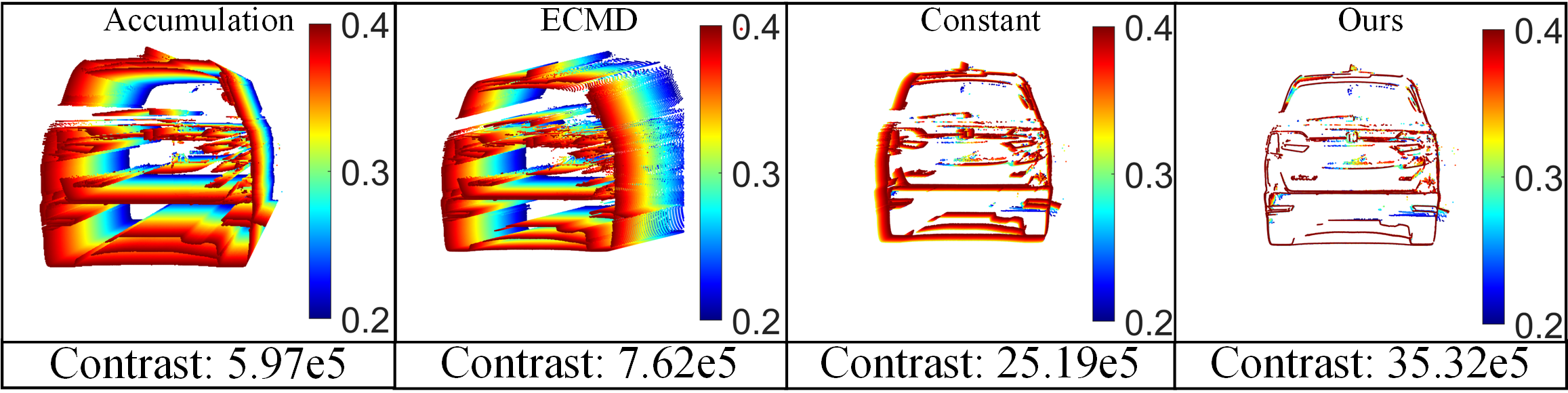}
    \end{center}
    \caption{Comparison of using different affine models.}
    \label{fig:warping_model}
\end{figure}

%% file: floats/fig_data_seq.tex
\begin{figure}[t!]
 \centering
 \includegraphics[width=\textwidth]{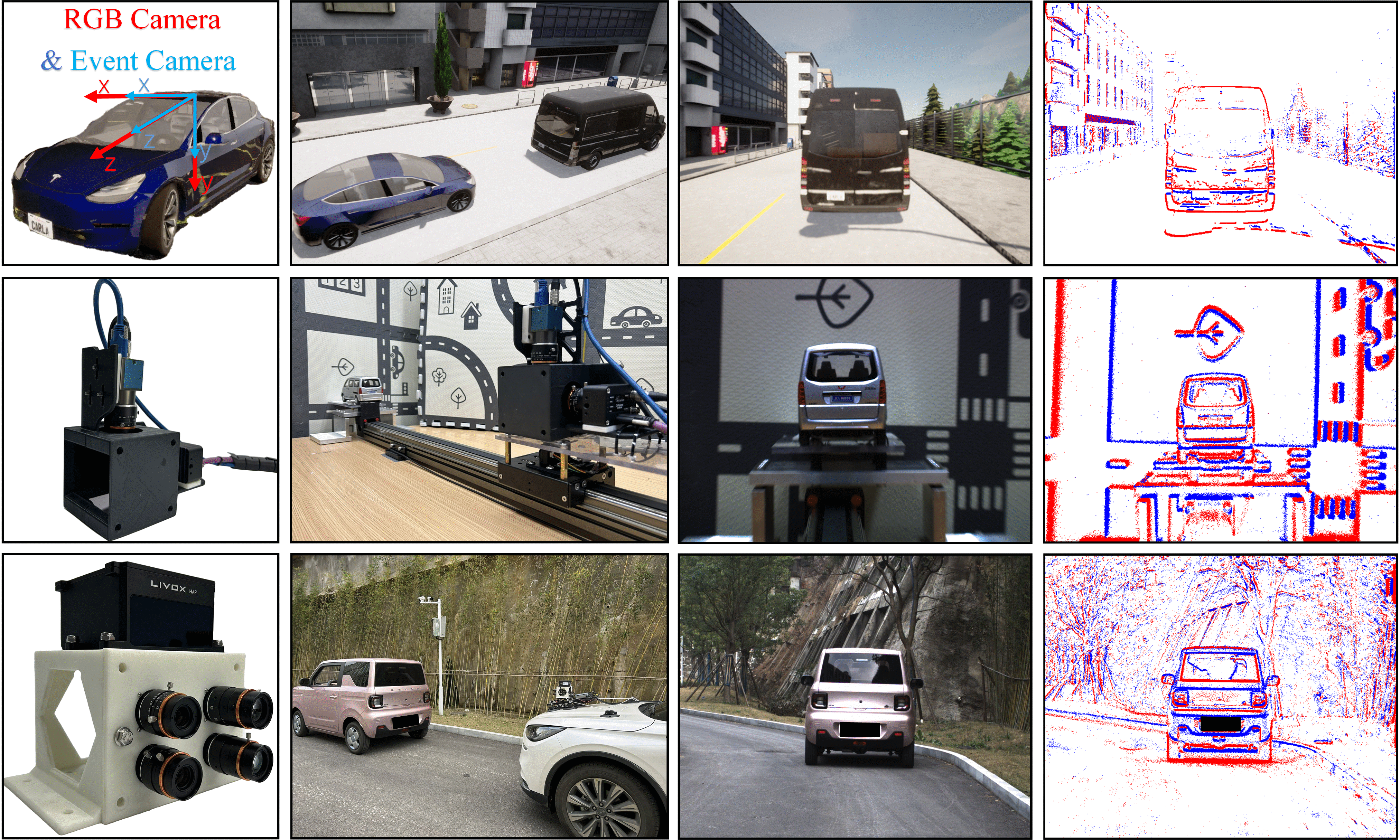}
 \caption{A snapshot of three datasets. 
 From top to bottom: \emph{Synthetic} dataset, \emph{Slider} dataset, and \emph{FCWD} dataset. 
 From left to right: platform configuration, third-person view of the TTC scenarios, intensity images, and event data (represented with a naive accumulation of events).}
\label{fig_data_seq}
\end{figure}

%% file: floats/table_hardware_specifications.tex
\begin{table}[t]
\centering
\caption{Hardware specifications for our datasets.} 
\begin{adjustbox}{max width=\linewidth}
\setlength{\tabcolsep}{3pt}
\begin{tabular}{c|cccc}
\toprule
Dataset Name   & 
Sensor Type &
Rate   &
Specifications &
Hardware-level Sync. \\
\midrule
\multirow{5}{*}{Synthetic}  & \multirow{2}{*}{Carla DVS}  & \multirow{2}{*}{N/A}  & $640\times480$ pixels                        & \multirow{2}{*}{-} \\
                            &                             &                       & FoV: \SI{52}{\degree}H / \SI{40}{\degree}V   &  \\
                            & \multirow{2}{*}{Carla RGB}  & \multirow{2}{*}{30Hz} & $640\times480$ pixels                        & \multirow{2}{*}{-} \\
                            &                             &                       & FoV: \SI{52}{\degree}H / \SI{40}{\degree}V   &  \\
                            & Carla Traffic Manager       & 1000Hz                & Report global location of all vehicle actors & - \\
\midrule
\multirow{7}{*}{Slider} & \multirow{2}{*}{Inivation DVXplorer}      & \multirow{2}{*}{N/A}  & $640\times480$ pixels                       & \multirow{2}{*}{\cmark} \\
                        &                                           &                       & FoV: \SI{20}{\degree}H / \SI{15}{\degree}V  &  \\
                        & \multirow{3}{*}{DAHENG MER2}              & \multirow{3}{*}{25Hz} & $1440\times1080$ pixels                     & \multirow{3}{*}{\cmark} \\
                        &                                           &                       & FoV: \SI{17}{\degree}H / \SI{13}{\degree}V  &  \\
                        &                                           &                       & color with global shutter                   &  \\
                        & \multirow{2}{*}{Encoder of slider motor}  & \multirow{2}{*}{100KHz} & Report the position and velocity of         & \multirow{2}{*}{\cmark} \\
                        &                                           &                         & the hybrid optical system on the slider.    &  \\
\midrule
\multirow{8}{*}{FCWD} 
        & 2$\times$Prophesee EVKv4                                                                              & \multirow{2}{*}{N/A}  & $1280\times720$ pixels        & \multirow{2}{*}{\cmark} \\
        & baseline: 7.5cm                                                                                       &                       & FoV: \SI{22}{\degree}H / \SI{12}{\degree}V  &  \\
        & \multirow{3}{*}{\begin{tabular}[c]{@{}c@{}} 2$\times$FLIR Blackfly S \\baseline: 7.5cm \end{tabular}} & \multirow{3}{*}{20Hz} & $1920\times1200$ pixels   & \multirow{3}{*}{\cmark} \\
        &                                                                                                       &                       & FoV: \SI{23}{\degree}H / \SI{15}{\degree}V  &  \\
        &                                                                                                       &                       & color with global shutter                   &  \\
        & \multirow{3}{*}{LiDAR: Livox HAP}  & \multirow{3}{*}{\begin{tabular}[c]{@{}c@{}} point rate: 452,000 points/s  \\frame rate: 10Hz \end{tabular}}  & range 150 m @ 10\% reflectivity               & \multirow{3}{*}{\cmark} \\
        &                             &                                    & FoV: \SI{120}{\degree}H / \SI{25}{\degree}V   &                         \\
        &                             &                                    & $\pm$3cm range precision @ 20 m               &                         \\

\bottomrule %
\end{tabular}
\end{adjustbox}
\label{tab:hardware_specification}
\end{table}

%% file: floats/fig_slider_demonstrate.tex
\begin{figure}[t!]
    \centering
    \begin{subfigure}[t]{0.51\linewidth}
        \centering
        \includegraphics[width=\textwidth]{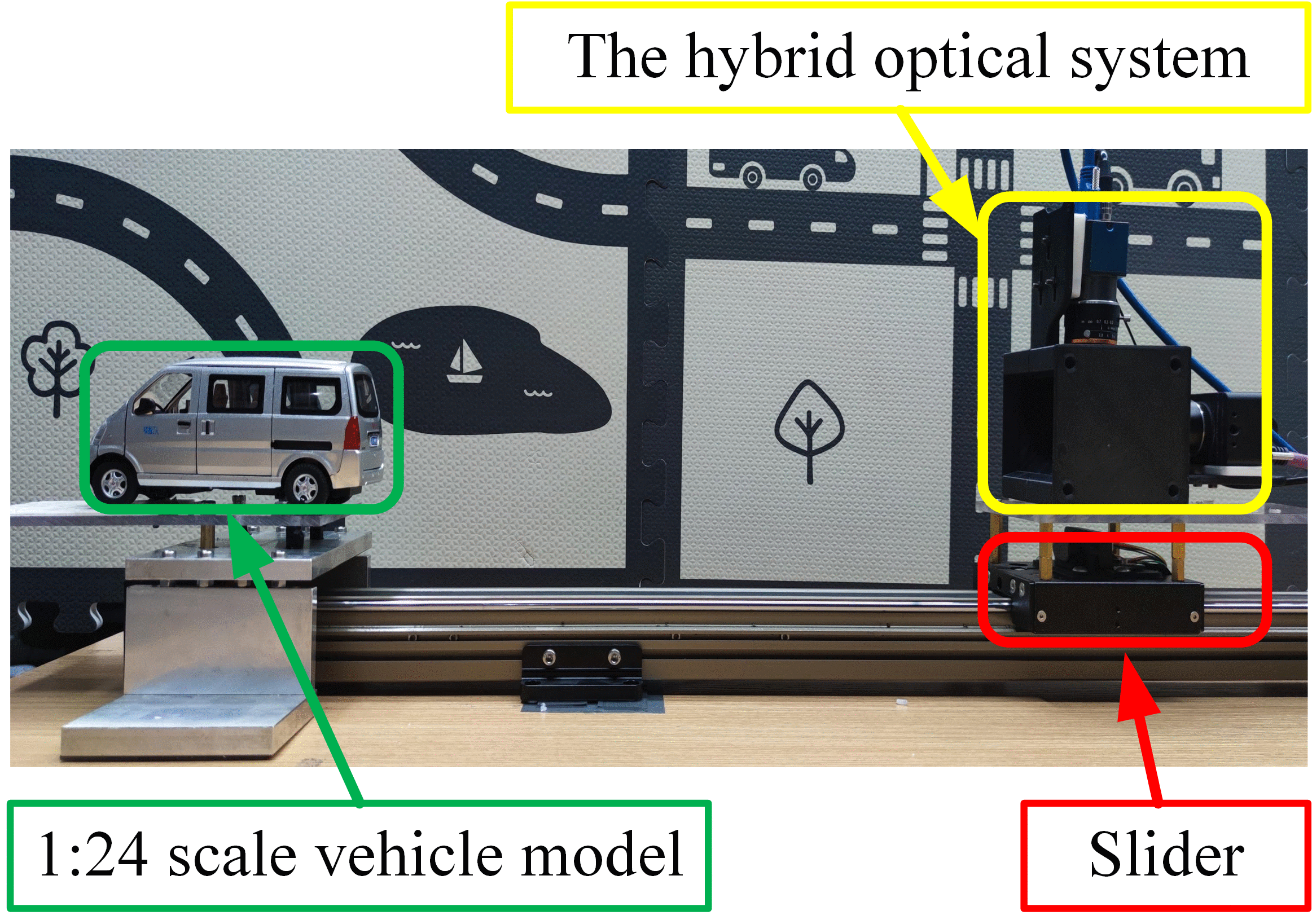}
        \caption{The proposed small-scale test platform.}
        \label{fig:small_scale_test_platform}
    \end{subfigure}
    \begin{subfigure}[t]{0.4\linewidth}
        \centering
        \includegraphics[width=\textwidth]{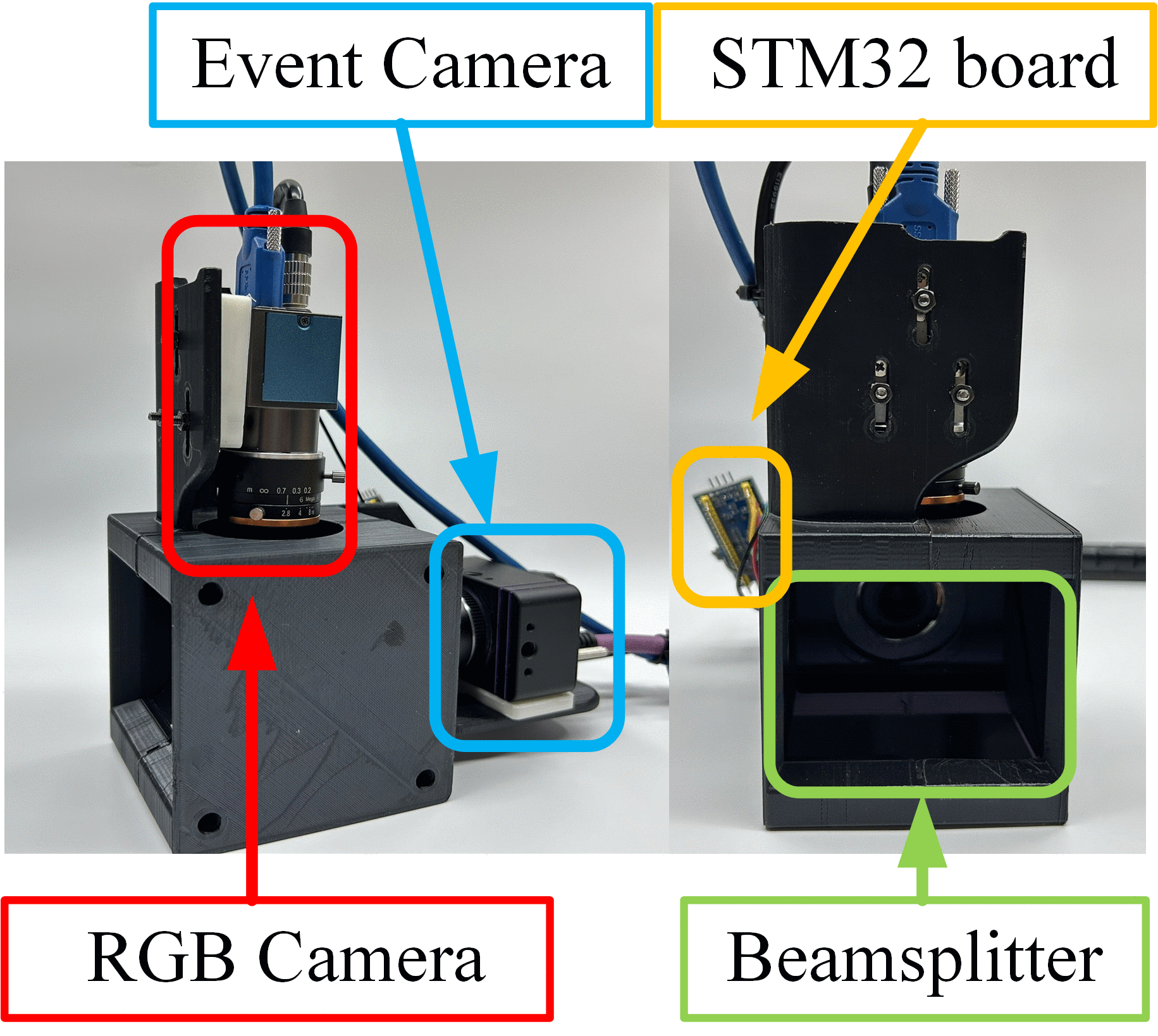}
        \caption{The composition of the hybrid optical system.}
        \label{fig:hybrid_optical_system}
    \end{subfigure}
    \caption{Illustration of our small-scale test platform and hybrid optical system.}
    \label{fig:slider_demonstrate}
\end{figure}

%% file: floats/table_dataset_comparision.tex
\begin{table}[t]
\centering
\caption{Comparison of Different Event-Centric Datasets.} 
\begin{adjustbox}{max width=\linewidth}
\setlength{\tabcolsep}{3pt}
\begin{tabular}{cccccccc}
\toprule
\multirow{2}{*}{Dataset} & Event      & \multirow{2}{*}{HFOV} & Urgent   & Event  & RGB     & \multirow{2}{*}{LiDAR}  & \multirow{2}{*}{Sync.}\\
                         & Resolution &                       & Brake   & Stereo & Stereo  &                         &                        \\
\midrule
MVSEC\cite{zhu2018multivehicle}  
& $346\times260$ 
& \SI{65}{\degree} 
& \xmark
& \cmark 
& \cmark 
& \cmark               
& Partially \\ %

DSEC\cite{gehrig2021dsec}           
& $640\times480$ 
& \SI{60}{\degree}
& \xmark
& \cmark 
& \cmark  
& \cmark                    
& Fully \\ %

ViViD++\cite{lee2022vivid++}          
& $640\times480$  
& \SI{44}{\degree}
& \xmark
& \xmark 
& \xmark 
& \cmark 
& Partially \\ %

M3ED\cite{Chaney_2023_CVPR}
& $1280\times720$
& \SI{63}{\degree}
& \xmark
& \cmark 
& \cmark 
& \cmark 
& Fully\\ %

Ours(FCWD)
& $1280\times720$
& \SI{22}{\degree}
& \cmark
& \cmark 
& \cmark 
& \cmark 
& Fully \\ %

\bottomrule %
\end{tabular}
\end{adjustbox}
\label{tab:dataset comparision}
\end{table}

%% file: floats/fig_strttc_experiment_result.tex
\begin{figure*}[t]
 \centering
 \includegraphics[width=\textwidth]{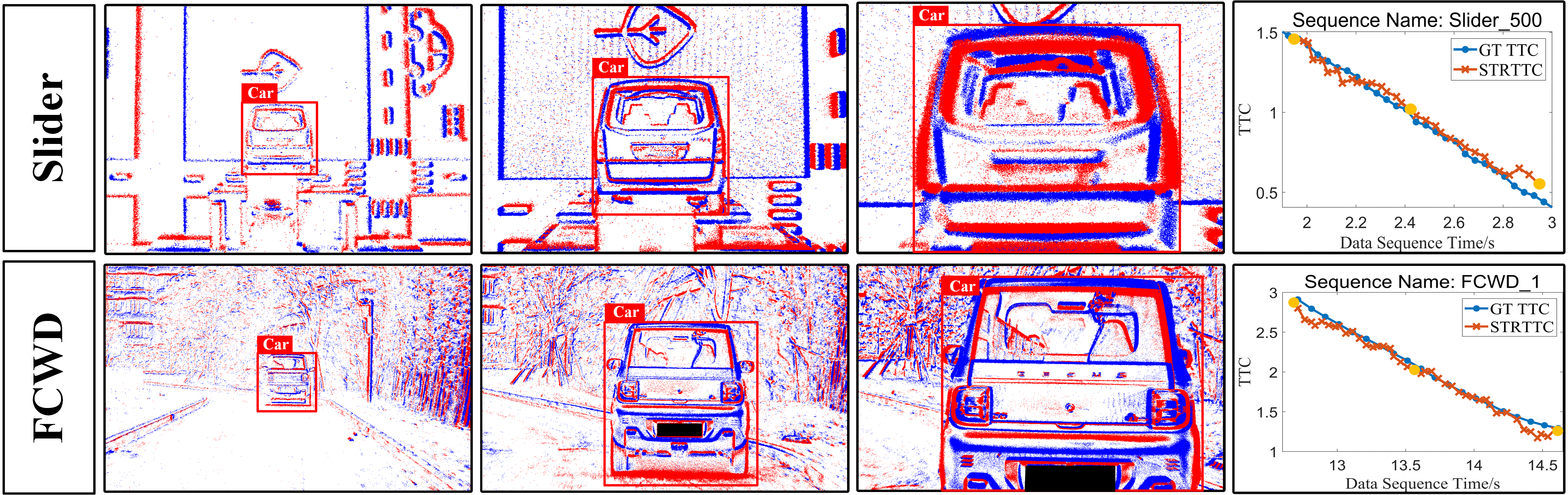}
 \caption{
 Illustration of the input (event data and the bounding box) and a continuous estimation results of TTC.
 From left to right: Three selected views of input data in a chronological order and the TTC estimation results through the whole process.
 Note that the first three columns correspond to the positions highlighted with yellow dots in the right-most plots.
 }
\label{fig_strttc_experiment_result}
\end{figure*}

%% file: floats/table_algorithm_comparison.tex
\begin{table}[t]
\centering
\caption{Quantitative Analysis of Each Method Under Different Parameters. Lower is better.}
\begin{adjustbox}{max width=\linewidth}
\setlength{\tabcolsep}{3pt}
\begin{tabular}{ccl|cc|cc|cc}
\toprule
& & \multirow{2}{*}{\textbf{Method}} & \multicolumn{2}{c|}{\emph{FCWD\_1}}  &   \multicolumn{2}{c|}{\emph{FCWD\_2}}  &   \multicolumn{2}{c}{\emph{FCWD\_3}} \\
\cline{4-9}
& &                                  &$e_{\text{TTC}}$     &Runtime (s)     &$e_{\text{TTC}}$     &Runtime (s)       &$e_{\text{TTC}}$     &Runtime (s)      \\
\midrule

& 
& Image's FoE~\cite{stabinger2016monocular}
& 5.39\%   & 0.017 
& 5.70\%   & 0.018
& 6.02\%   & 0.017 \\
\addlinespace

& 
& FAITH~\cite{dinaux2021faith}
& 25.49\%    & 0.214 
& 27.91\%    & 0.203 
& 39.15\%    & 0.210 \\
\addlinespace

\multirow{6}{*}{\rotatebox[origin=c]{90}{Event Num}} 
& \multirow{2}{*}{\rotatebox[origin=c]{90}{$1e5$}}
& CMax\cite{Gallego18cvpr}
& 5.68\%    & \underline{1.73}
& 5.65\%    & \underline{1.93}
& 6.92\%    & \underline{1.79}\\
& 
& Our Init + CMax~\cite{Gallego18cvpr}
& 5.85\%    & \underline{1.39}
& 4.05\%    & \underline{1.61}
& 4.52\%    & \underline{1.47}\\

& \multirow{2}{*}{\rotatebox[origin=c]{90}{$2e5$}}
& CMax\cite{Gallego18cvpr}
& \underline{2.42\%}  &  2.53
& \underline{2.39\%}  &  3.12
& \underline{3.04\%}  &  3.34\\
& 
& Our Init + CMax~\cite{Gallego18cvpr}
&  2.33\%  &  1.94
&  2.26\%  &  2.38
&  2.97\%  &  2.59\\

& \multirow{2}{*}{\rotatebox[origin=c]{90}{$3e5$}}
& CMax\cite{Gallego18cvpr}
& 6.04\%    & 2.93
& 4.37\%    & 2.83
& 3.37\%    & 2.83\\
& 
& Our Init + CMax~\cite{Gallego18cvpr}
& \underline{2.10\%}    & 2.13
& \underline{2.12\%}    & 1.99
& \underline{2.57\%}    & 2.14\\
\addlinespace

\multirow{3}{*}{\rotatebox[origin=c]{90}{Neighboring Size}} 
& \multirow{3}{*}{\rotatebox[origin=c]{90}{$s=2$}}
& ETTCM Scaling~\cite{nunesTimeToContact2023}
& \underline{15.52\%} & \underline{0.307}
& 18.45\% & \underline{0.282}
& \underline{19.08\%} & \underline{0.295}\\
&
& ETTCM Translation~\cite{nunesTimeToContact2023}
& 19.44\%    & 0.839
& \underline{17.30\%}    & 0.772
& 21.20\%    & 0.806\\
&
& ETTCM 6-DOF~\cite{nunesTimeToContact2023}
& 28.00\%    & 1.09
& 27.44\%    & 0.997
& 36.78\%    & 1.05\\
& \multirow{3}{*}{\rotatebox[origin=c]{90}{$s=3$}}
& ETTCM Scaling~\cite{nunesTimeToContact2023}
& 28.07\%    & 0.526
& 32.04\%    & 0.484
& 266.50\%   & 0.458\\
&
& ETTCM Translation~\cite{nunesTimeToContact2023}
& 124.72\%    & 1.53
& 117.06\%    & 1.41
& 109.84\%    & 1.44\\
&
& ETTCM 6-DOF~\cite{nunesTimeToContact2023}
& 25.23\%    & 2.02
& 32.52\%    & 1.82
& 36.43\%    & 1.91\\
\addlinespace

&
& Ours
& 9.84\%    &   0.017
& 11.55\%   &   0.023
& 14.09\%   &  0.025 \\

\bottomrule %
\end{tabular}
\end{adjustbox}
\label{tab:algorithm_comparison}
\end{table}

%% file: floats/fig_runtime_accuracy_comparison.tex
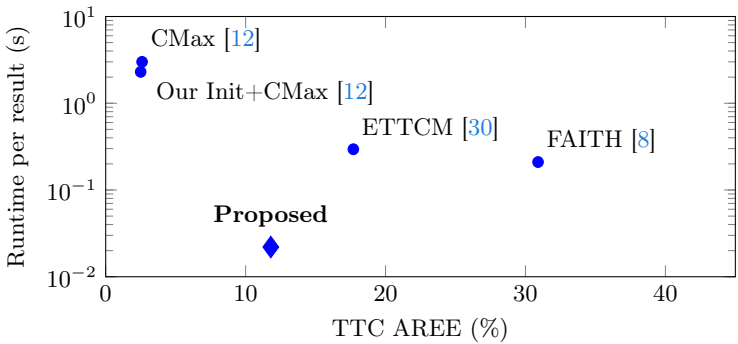
\begin{figure}[t!]
\centering
\begin{tikzpicture}

\begin{axis}[
    width=10cm, %
    height=5cm, %
    xlabel={TTC AREE (\%)},
    ylabel={Runtime per result (s)},
    xmin=0, xmax=45,
    ymin=0.01, ymax=10,
    xmode=normal,
    ymode=log,
    legend style={
        at={(0.5,1.03)}, %
        anchor=south,
        legend columns=-1, %
        /tikz/every even column/.append style={
            column sep=2mm %
        }
    },
    legend cell align={left},
    axis lines=box,
]

\addplot[
    only marks,
    color=blue,
    mark=*,
    mark size=2pt
] coordinates {
    (2.6,3)
    (2.5,2.3)
    (30.9,0.21)
    (17.7,0.295)
};

\addplot[
    only marks,
    color=blue,
    mark=diamond*,
    mark size=4pt
] coordinates {
    (11.8,0.022)
};

\node at (axis cs:2.6,3.1) [anchor=south west] {CMax\cite{Gallego18cvpr}};
\node at (axis cs:3,2.3) [anchor=north west] {Our Init+CMax~\cite{Gallego18cvpr}};
\node at (axis cs:30.9,0.21) [anchor=south west] {FAITH~\cite{dinaux2021faith}};
\node at (axis cs:17.7,0.29) [anchor=south west] {ETTCM~\cite{nunesTimeToContact2023}};
\node at (axis cs:11.8,0.03) [anchor=south] {\textbf{Proposed}};

\end{axis}
\end{tikzpicture}

\caption{\textbf{Runtime and Accuracy Comparison} of event based TTC Estimation Methods: An Average Performance Evaluation on our FCWD.}
\label{fig:runtime_accuracy}
\end{figure}